\documentclass[a4paper,fleqn]{cas-sc}

\usepackage[numbers]{natbib}
\usepackage{amsmath}
\usepackage{amssymb}
\usepackage{array}
\usepackage{booktabs}
\usepackage{caption}
\usepackage{graphicx}
\usepackage{hyperref}
\usepackage{lastpage}
\usepackage{multirow}
\usepackage{tikz}
\usepackage{xcolor}
\usetikzlibrary{arrows.meta,positioning,calc,fit,backgrounds,shapes.geometric,shapes.misc}

\newcommand{\method}{Warrant}

\newcounter{algorithm}

\begin{document}
\renewcommand{\lastpage}{\pageref*{LastPage}}

\let\WriteBookmarks\relax
\def\floatpagepagefraction{0.7}
\def\textpagefraction{.01}

\shorttitle{Warranted Attention}
\shortauthors{Yu and Ha}

\title[mode=title]{Warranted Attention: Learning What to Pass from Attention to Prediction}

\author[1]{Minwoo Yu}
\ead{snowypainter@konkuk.ac.kr}
\credit{Conceptualization, Methodology, Software, Validation, Investigation, Writing - original draft}

\author[1]{Young-guk Ha}
\cormark[1]
\ead{ygha@konkuk.ac.kr}
\credit{Supervision, Project administration, Funding acquisition, Writing - review \& editing}

\cortext[cor1]{Corresponding author}

\affiliation[1]{organization={Smart Computing Laboratory, Department of Computer Science \& Engineering, Konkuk University},
            addressline={120 Neungdong-ro, Gwangjin-gu},
            city={Seoul},
            postcode={05029},
            country={Republic of Korea}}

\begin{abstract}
Relevance of information read by attention does not guarantee that its contribution benefits the current prediction. We propose Warrant, which learns how strongly attention-derived item contributions should be transmitted under the current query. Warrant applies learned item-wise permission before aggregation without renormalization, jointly controlling relative allocation and total transmission mass. Across backbones, we construct interfaces connecting these contributions to prediction scores or states and compare against ungated models on the same paths. On three CyGNet datasets, ungated paths reduce MRR, whereas learned permission partially or almost fully recovers the losses. In a 5-seed HotpotQA/RoBERTa experiment, distractors receive lower permission than gold support, while mean Support MRR rises from .9111 to .9141 and the unsupported selection rate falls from .2271 to .2215. Performance comparisons and contribution-level interventions across five task families reveal both the effects and limits of selective control. The results support learning the strength of contributions transmitted to prediction separately from attention relevance.
\end{abstract}

\begin{highlights}
\item Warrant learns permission for attention-derived item contributions.
\item Same-path controls separate learned permission from the effect of added paths.
\item Permission recovers losses caused by ungated contributions in CyGNet.
\item Learned gates suppress distractors and improve HotpotQA evidence selection.
\end{highlights}


\maketitle

\section{Introduction}
\label{sec:introduction}

Attention computes compatibility between a current query and key--value items to determine how much information to read from each item. However, information relevant to a query need not benefit the current prediction. A passage resembling a question may not support its answer, and a historical fact associated with a query may reinforce a previously frequent candidate rather than the current correct one. Prior interpretability studies have also observed differences between attention weights and actual contributions \cite{jain2019attention,serrano2019attention,kobayashi2020attention}. We study how to control the effects of this mismatch on prediction.

Separate paths connecting memory or historical information to prediction are used across several fields. Neural Cache and kNN language models mix memory-derived distributions with existing predictive distributions, while pointer-generator networks directly incorporate source attention mass into output-token probabilities \cite{grave2017cache,khandelwal2020knnlm,see2017get}. Tip-Adapter combines predictions read from a key--value cache with an existing classifier, and CyGNet mixes historical-copy and generation predictions \cite{zhang2022tipadapter,zhu2021cygnet}. These paths provide additional information, but not every transmitted contribution is useful. Excessive reliance on history and noisy retrieval can degrade predictions, motivating adaptive retrieval and confidence-based control \cite{zhu2021cygnet,zheng2021adaptiveknn,jiang2022robustknn}.

Our question is \emph{how strongly each item contribution read by attention should be transmitted to prediction under the current query}. We propose \method{}, which learns this transmission strength as permission. Before aggregation, \method{} applies a scalar gate taking both the query and the item as inputs to each attention-weighted value term. Without renormalizing the resulting weights, it can change both the relative allocation across items and the total transmission mass of the path. Permission is the transmission strength of a contribution learned through the task objective.

Different backbones make predictions through edge scores, mark logits, evidence scores, location states, or tail scores. We connect attention-derived contributions to these prediction interfaces and apply the same permission operation. Base, OpenPath, and Full distinguish the effects of path construction from those of learned control. OpenPath uses exactly the same path as Full but fixes the gate to 1. Comparing Full with OpenPath therefore measures the additional effect of learning permission on that path.

Ungated paths can indeed be harmful. For CyGNet on GDELT, ICEWS14, and ICEWS18, OpenPath reduces MRR relative to Base by .0025, .0093, and .0098, respectively, while learned permission recovers .0025, .0094, and .0063. In a 5-seed HotpotQA/RoBERTa analysis, the mean gate for gold support is .8882, compared with .5054/.4665 for hard/random distractors. In the same experiment, Support MRR increases from .9111 for OpenPath to .9141 for Full, while the unsupported selection rate decreases from .2271 to .2215. These cases show selective contribution control occurring alongside changes in prediction.

We evaluate applicability across five task families: CTDG, MTPP, RAG, STPP, and TKG. We report both overall Base--Full comparisons and same-path OpenPath--Full comparisons, and analyze alternative control mechanisms using scalar, item-only, normalized, attention-adapter, and G1/G2 controls. Item-removal and evidence-mass interventions also measure whether harmful contributions are present. Permission effects vary by task and backbone, with small or negative increments in some settings.

Our contributions are as follows.
\begin{itemize}
\item We formulate a prediction-control problem that learns the transmission strength of attention-derived item contributions and present \method{} to apply permission to these contributions.
\item We analyze the relationship between contribution control and prediction changes through harmful-path recovery in CyGNet and learned distractor suppression in HotpotQA.
\item We evaluate the effects and applicability limits of permission through same-path comparisons, alternative controls, and contribution interventions across five task families.
\end{itemize}

\section{Related Work}
\label{sec:related_work}

\subsection{Attention and Prediction Contribution}
\label{subsec:rw_attention_memory}

Bahdanau attention lets a decoder read relevant parts of a source sentence, and the Transformer aggregates values according to query--key compatibility \cite{bahdanau2015neural,vaswani2017attention}. End-to-End Memory Networks likewise read memory in response to a query \cite{sukhbaatar2015end}. In these approaches, a high attention weight indicates the proportion read from an item, rather than how strongly the resulting information supports a particular prediction.

Jain and Wallace, Serrano and Smith, and Wiegreffe and Pinter analyze the interpretation of attention weights as explanations \cite{jain2019attention,serrano2019attention,wiegreffe2019attention}. Kobayashi et al. analyze contributions while accounting for transformed value magnitudes, and Value Zeroing measures context mixing by removing value information \cite{kobayashi2020attention,mohebbi2023quantifying}. We address a control problem motivated by these observations: learning permission over already formed item contributions and evaluating its effect on a shared prediction path.

\subsection{Direct Memory-to-Prediction Paths}
\label{subsec:rw_direct_paths}

Separate connections from memory to prediction appear in several tasks. Neural Cache and kNN-LM mix memory-derived token distributions with language-model distributions \cite{grave2017cache,khandelwal2020knnlm}. Pointer-generator networks connect source attention mass to output-token probabilities \cite{see2017get}, and Tip-Adapter combines key--value cache predictions with a CLIP classifier \cite{zhang2022tipadapter}. CyGNet combines historical-copy and generation distributions \cite{zhu2021cygnet}. These methods establish precedents for connecting memory-derived contributions to prediction.

Other studies control the quality of information carried by additional paths. Adaptive kNN-MT adaptively selects among noisy neighbors, while Robust kNN-MT addresses retrieval noise \cite{zheng2021adaptiveknn,jiang2022robustknn}. CyGNet's history--generation mixture and CENET's query-dependent historical modeling also avoid treating all historical information uniformly \cite{zhu2021cygnet,xu2023cenet}. \method{} controls attention-derived item contributions by applying permission based on the query and item before aggregation. Path construction connects this operation to different backbones, and OpenPath provides a control for measuring learned permission on the same connection.

\subsection{Gating and Conditional Contribution Control}
\label{subsec:rw_gating_modulation}
\label{subsec:rw_gated_attention}

Highway Networks, recurrent gates, and GLUs control the transmission of transformed features, memory states, or activations \cite{srivastava2015highway,hochreiter1997long,cho2014gru,dauphin2017language}. FiLM modulates features using a conditioning signal, squeeze-and-excitation reweights channels, and MoE learns expert routing \cite{perez2018film,hu2018squeeze,shazeer2017outrageously}. Gated-Attention Reader, GTrXL, and Mega also use gates for query-specific reading, training stabilization, and long-sequence modeling, respectively \cite{dhingra2017gated,parisotto2020gated,ma2022mega}.

Item-wise control appears in Residual Gated Graph ConvNets and DIN; DIN weights history without a softmax sum-to-one constraint \cite{bresson2017residual,zhou2018din}. Sigmoid attention studies the theory and stabilization of non-normalized weighting \cite{ramapuram2025sigmoid}. These methods employ query-dependent item-wise control and non-normalized weighting. We use such operations to control how strongly attention-weighted contributions reach prediction, analyzing harmful-path recovery and distractor suppression with same-path controls.

Edge-augmented Graph Transformer (EGT) constructs node-pair representations from graph structural matrices and edge features, updating them at each layer to compute attention biases and sigmoid gates \cite{hussain2022egt}. EGT multiplies softmax weights by pair-specific gates and aggregates values without renormalization, sharing this aggregation structure with \method{}. Its gates use updated pair representations rather than only source-item inputs. EGT is a graph architecture that jointly learns node and edge channels; \method{} computes permission from a backbone's queries and keys to control contributions at task-specific prediction interfaces.

EGT also evaluates an ungated variant to measure the effect of gates within its graph architecture \cite{hussain2022egt}. We compare OpenPath and Full on a shared prediction path and combine contribution removal with labeled-distractor analysis to assess selective transmission. The studies differ in gate input representations, architectural integration, and the question evaluated, while sharing post-softmax multiplication.

Another closely related attention configuration is Qiu et al.'s G1/G2 \cite{qiu2025gated}. G2 gates value projections using the hidden state of each key/value token before aggregation. G1 gates the aggregate SDPA output using the current query token's hidden state. \method{} controls each weighted term using the current query--item pair before aggregation. Applying a scalar gate to a value before weighting or to the weighted value afterward is algebraically equivalent; the relevant distinctions are conditioning, preservation of the item axis, and the prediction interface. Table~\ref{tab:gated_object_taxonomy} summarizes the control targets, and Section~\ref{subsec:qiu_gated_attention_comparison} reports direct experiments with G1/G2.

\begin{center}
\begin{minipage}{\textwidth}
\centering
\captionof{table}{Control targets and units of representative methods. Qiu G2 computes value gates from items, while G1 computes aggregate gates from queries. \method{} uses both query and item to compute per-contribution permission. Multiplying a scalar gate into the value or the weighted value is equivalent. The normalization column refers to the representative configurations listed.}
\label{tab:gated_object_taxonomy}
\scriptsize
\setlength{\tabcolsep}{3.2pt}
\renewcommand{\arraystretch}{1.12}
\resizebox{\textwidth}{!}{
\begin{tabular}{lllll}
\toprule
Family & Gated object & Gate unit & Attention renormalized & Difference from \method{} \\
\midrule
Highway / recurrent & layer, carry, or memory update & vector / channel & no & layer-level flow \\
GLU / gated FFN & hidden activation & token / channel & no & activation-level gate \\
SE / channel attention & feature channel & channel & no & channel-level gate \\
Attention-logit / sparse attention & relevance score or support & item logit / item set & yes & changes \(\alpha_{ij}\) and redistributes mass \\
Normalized MoE gate & expert output mixture & expert & mixture-normalized & expert-routing gate \\
Copy / pointer gate & generation-copy branch & branch / candidate & branch-normalized & task branch gate \\
Qiu G2: value gate & \(v_j\) before attention sum & item / channel & no & item input; before item mixing \\
Qiu G1: SDPA-output gate & \(h_i=\sum_j\alpha_{ij}v_j\) & query / head / channel & no & query input; after item mixing \\
EGT \cite{hussain2022egt} & attention-weighted node value & node pair / head & no & evolving edge-representation input \\
Post-attention gate & \(h_i=\sum_j\alpha_{ij}v_j\) & aggregate vector & no & item identity is already mixed \\
\method{} & \(\alpha_{ij}v_j\) & query-item weighted value term & no & gates metric-facing contribution directly \\
\bottomrule
\end{tabular}
}
\end{minipage}
\end{center}

\subsection{Evidence Selection and Prediction Calibration}
\label{subsec:rw_retrieval_calibration}
\label{subsec:rw_positioning}

Dense retrieval retrieves candidate passages, and RAG and FiD integrate retrieved information into generation \cite{karpukhin2020dense,lewis2020rag,izacard2021leveraging}. Longformer supports long-context processing, while passage reranking and evidentiality-guided generation address candidate ordering and evidential roles \cite{beltagy2020longformer,nogueira2019passage,asai2022evidentiality}. Calibration and selective prediction adjust output confidence or reject uncertain predictions \cite{guo2017calibration,geifman2017selective}. Our RAG evaluation is limited to supporting evidence selection. Permission controls internal contributions from the selected candidates to evidence ranking.

\section{Problem Formulation}
\label{sec:problem_definition}

\subsection{Attention-Derived Contributions}
Let \(q_i\) denote the query for the current prediction request, and let \((k_j,v_j)\) denote an item's key and value. Attention normalizes compatibility scores to produce item-wise access weights. The resulting weighted values form the aggregate in Eq.~\eqref{eq:base_attention}.
\begin{equation}
\label{eq:base_attention}
\alpha_{ij}=\operatorname{softmax}_j s(q_i,k_j),
\qquad h_i=\sum_j\alpha_{ij}v_j.
\end{equation}
The control target for each item is defined in Eq.~\eqref{eq:weighted_value_contribution}.
\begin{equation}
\label{eq:weighted_value_contribution}
c_{ij}=\alpha_{ij}v_j.
\end{equation}
The contribution in Eq.~\eqref{eq:weighted_value_contribution} is a computational term. Whether it supports the correct prediction depends on its value direction, other items, and the downstream predictor. A high attention weight alone therefore does not imply positive prediction utility.

\subsection{Permission}
We learn each contribution's transmission strength as a function of the current query and item. Equation~\eqref{eq:joint_warrant} defines permission and the controlled aggregate.
\begin{equation}
\label{eq:joint_warrant}
g_{ij}=g(q_i,k_j),\qquad
h_i^W=\sum_j g_{ij}c_{ij},\qquad
g_{ij}\in[\lambda,1].
\end{equation}
In Eq.~\eqref{eq:joint_warrant}, \(\lambda>0\) is the minimum transmission ratio. Control is applied while the item axis is preserved, and permission is learned through the downstream task loss.

\subsection{Relative Allocation and Total Transmission}
Attention weights sum to 1, whereas the sum of effective weights \(\alpha_{ij}g_{ij}\) can vary. Equation~\eqref{eq:permission_mass} defines the total transmission factor and its corresponding normalized distribution.
\begin{equation}
\label{eq:permission_mass}
m_i=\sum_j\alpha_{ij}g_{ij},\qquad
\bar\alpha_{ij}=\frac{\alpha_{ij}g_{ij}}{m_i},
\qquad h_i^W=m_i\sum_j\bar\alpha_{ij}v_j.
\end{equation}
Equation~\eqref{eq:permission_mass} shows that permission controls both relative allocation and total transmission mass. Combining normalized reweighting with a separate scalar \(m_i\) represents the same effective coefficients. The distinction is therefore between control limited to normalization and control that also changes total mass. Here, \(m_i\) is the sum of effective weights.

\subsection{Prediction-Facing Interfaces}
The prediction object \(z_i\) is an edge score, candidate-mark logit, support score, next-location state, or tail score. Equation~\eqref{eq:prediction_interface} represents a prediction interface receiving the backbone's existing state and the controlled aggregate.
\begin{equation}
\label{eq:prediction_interface}
z_i=F_\theta(x_i,h_i^W).
\end{equation}
The specific connection in Eq.~\eqref{eq:prediction_interface} varies by task. Permission can be applied to an existing aggregation, or an additional path can transmit item-derived contributions to a score or state. Section~\ref{subsec:domain_adaptation} and Appendix~\ref{subsec:path_localization} describe this implementation procedure. Table~\ref{tab:task_adaptation} summarizes the items and prediction targets for each task.

\begin{center}
\begin{minipage}{\textwidth}
\centering
\captionof{table}{Items and prediction interfaces by task. The same permission mechanism controls item contributions connected to each backbone's score or state.}
\label{tab:task_adaptation}
\resizebox{\textwidth}{!}{
\begin{tabular}{llll}
\toprule
Domain & Key-value item & Prediction target & Metric-defining value path \\
\midrule
CTDG & temporal neighbor / interaction history & current source-destination edge & edge score \\
MTPP & past marked event & next event mark & candidate mark logit \\
RAG & retrieved passage & supporting evidence passage & support ranking logit \\
STPP & past spatio-temporal event & next event location & next-location dynamics state \\
TKG & historical temporal fact & tail entity & tail ranking score / copy path \\
\bottomrule
\end{tabular}
}
\end{minipage}
\end{center}

\subsection{Evaluation on a Shared Path}
Base is the original model, OpenPath fixes \(g=1\) on the constructed path, and Full learns permission on that same path. GenericGate is a separate placement control that adds a gate at a generic attention location in the backbone. For a higher-is-better metric \(M\), Eq.~\eqref{eq:component_estimands} defines the evaluation quantities.
\begin{align}
\label{eq:component_estimands}
\Delta_{\mathrm{path}}&=M(\mathrm{OpenPath})-M(\mathrm{Base}),\\
\Delta_{\mathrm{perm}}&=M(\mathrm{Full})-M(\mathrm{OpenPath}).
\end{align}
The signs in Eq.~\eqref{eq:component_estimands} are reversed for lower-is-better metrics. Full--Base measures the change in the adapted system, including path construction, whereas \(\Delta_{\mathrm{perm}}\) measures the additional change from learned permission on the same path. Because each variant is retrained, these differences also include representation adaptation during learning.

\section{Method}
\label{sec:method}

\method{} learns how strongly each item contribution read by attention is transmitted under the current query. Control is applied before items are aggregated, and the controlled contributions are passed to the backbone's prediction interface.

\subsection{Learning Permission}
\label{subsec:warrant_operator}

The item-wise term formed by an attention weight and a value is the unit of control. In Eq.~\eqref{eq:warrant_value_term}, $i$ indexes queries and $j$ indexes items.
\begin{equation}
\label{eq:warrant_value_term}
c_{ij}=\alpha_{ij}v_j.
\end{equation}
A scalar score is computed from the current query and the item's key. The parameters in Eq.~\eqref{eq:warrant_score} are learned through the task objective.
\begin{equation}
\label{eq:warrant_score}
\psi(q_i,k_j)=w^\top\mathrm{GELU}(W_q q_i+W_k k_j)+b.
\end{equation}
This score is transformed into bounded permission. The leak factor $\lambda$ in Eq.~\eqref{eq:warrant_gate} prevents valid-item connections from becoming hard zeros; its default value is $0.05$.
\begin{equation}
\label{eq:warrant_gate}
g_{ij}=\lambda+(1-\lambda)\sigma(\psi(q_i,k_j)).
\end{equation}
Each item contributes to prediction through the permission-scaled term in Eq.~\eqref{eq:warrant_value_term_gate}.
\begin{equation}
\label{eq:warrant_value_term_gate}
c_{ij}^{W}=g_{ij}c_{ij}=\alpha_{ij}g_{ij}v_j.
\end{equation}
Summing the controlled terms gives the representation in Eq.~\eqref{eq:warrant_aggregation}. Effective weights are not renormalized, allowing both relative allocation and the path's total weight mass to vary.
\begin{equation}
\label{eq:warrant_aggregation}
h_i^W=\sum_j c_{ij}^{W}.
\end{equation}
In Eq.~\eqref{eq:warrant_aggregation}, equal gates scale the entire path, whereas item-specific gates also change relative contributions. Total weight mass is determined by the sum of coefficients, while output-vector magnitude depends on the combination of value directions. Figure~\ref{fig:warrant_operator} illustrates this item-wise operation.

\begin{center}
\resizebox{\textwidth}{!}{%
\begin{tikzpicture}[
    font=\sffamily\normalsize,
    >=Latex,
    box/.style={
        draw=black!55,
        thick,
        rounded corners=5pt,
        align=center,
        minimum height=12mm,
        minimum width=32mm
    },
    inputbox/.style={box, fill=gray!10},
    gatebox/.style={box, fill=blue!8, draw=blue!60, minimum width=34mm},
    warrantbox/.style={box, fill=teal!12, minimum width=36mm},
    outputbox/.style={box, fill=green!10, minimum width=48mm},
    op/.style={
        draw=black!60,
        thick,
        circle,
        fill=white,
        minimum size=8mm,
        inner sep=0pt
    },
    labelstyle/.style={
        font=\sffamily\bfseries\scriptsize,
        align=center,
        text=black!70
    },
    warrlabel/.style={
        font=\sffamily\bfseries\scriptsize,
        align=center,
        text=blue!70
    },
    arrow/.style={->, very thick, draw=black!70},
    warrarrow/.style={->, very thick, draw=blue!70}
]

\node[inputbox] (base) at (0,0) {\(\alpha_{ij}v_j\)};
\node[op] (mul) at (3.0,0) {\(\times\)};
\node[warrantbox] (wterm) at (5.7,0) {\(\alpha_{ij}g_{ij}v_j\)};
\node[outputbox] (state) at (10.7,0) {\(\displaystyle h_i^W=\sum_j \alpha_{ij}g_{ij}v_j\)};

\node[gatebox] (gate) at (3.0,1.8) {\(g_{ij}=g(q_i,k_j)\)};

\draw[arrow] (base.east) -- (mul.west);
\draw[warrarrow] (gate.south) -- node[right, font=\sffamily\scriptsize, text=blue!70] {scaling} (mul.north);
\draw[arrow] (mul.east) -- (wterm.west);
\draw[arrow] (wterm.east) -- node[below, font=\sffamily\scriptsize, yshift=-1pt] {aggregate over \(j\)} (state.west);

\node[labelstyle] at (0,-0.95) {attention-weighted\\value term};
\node[warrlabel] at (3.0,2.75) {value-contribution\\interface};
\node[warrlabel] at (5.7,-0.95) {warranted\\term};
\node[labelstyle] at (10.7,-0.95) {prediction\\state};

\end{tikzpicture}
}
\captionof{figure}{Permission before aggregation. Item-wise permission is applied to each attention-weighted value term before aggregation. The controlled contributions are transmitted to the backbone's prediction interface.}
\label{fig:warrant_operator}
\end{center}

\subsection{Control Before Aggregation}

\paragraph{Value and weighted-value control.}
For the same scalar gate and attention weight, gating the value before weighting is equivalent to gating the weighted value. Equation~\eqref{eq:value_gate_equivalence} shows why conditioning and item-wise control, rather than multiplication order, are the relevant properties.
\begin{equation}
\label{eq:value_gate_equivalence}
\alpha_{ij}(g_{ij}v_j)=g_{ij}(\alpha_{ij}v_j).
\end{equation}
The same item receives different $g_{ij}$ values for different queries. Qiu et al.'s G2 \cite{qiu2025gated} computes a value gate from the key/value token's hidden state, while G1 uses the query token's hidden state to gate the output after aggregation. \method{} uses both the query and the item to control each term before aggregation. Section~\ref{subsec:qiu_gated_attention_comparison} compares these configurations.

\paragraph{Attention logits and transmission mass.}
Modifying attention logits followed by softmax leaves the weights summing to 1. In contrast, Eq.~\eqref{eq:warrant_aggregation} does not force mass suppressed from one item to be redistributed to others. Warrant retains an explicit total transmission factor. For positive gates and original attention logits $s_{ij}$, Eq.~\eqref{eq:logit_mass_equivalence} expresses the same computation as normalized reweighting combined with a separate mass factor.
\begin{equation}
\label{eq:logit_mass_equivalence}
\begin{aligned}
m_i&=\sum_j\alpha_{ij}g_{ij},\\
\alpha_{ij}g_{ij}
&=m_i\,\operatorname{softmax}_{j}(s_{ij}+\log g_{ij}).
\end{aligned}
\end{equation}
Omitting $m_i$ in Eq.~\eqref{eq:logit_mass_equivalence} leaves only relative allocation. Normalized, scalar, and attention-adapter controls compare the effects of these degrees of freedom (Section~\ref{subsec:permission_function_controls}).

\paragraph{Item contributions and prediction logits.}
Attention logits determine the proportions read from items, whereas prediction logits score the final candidates. \method{} gates item contributions, whose outputs can then be connected to prediction logits or states. Adding controlled historical mass to a tail score in TKG differs from gating the entire final tail logit. The latter jointly scales contributions already combined into one score, whereas the former controls individual items contributing to the same candidate.

Moving a gate across a linear projection preserves the result. With bias treated as a separate term, Eq.~\eqref{eq:linear_gate_projection} holds.
\begin{equation}
\label{eq:linear_gate_projection}
W\left(\sum_j g_{ij}c_{ij}\right)=\sum_j g_{ij}Wc_{ij}.
\end{equation}
Equation~\eqref{eq:linear_gate_projection} does not directly extend across nonlinear activations or normalization. The control unit is therefore determined by whether an item-wise decomposition is preserved, rather than whether a tensor is named a value or a score. A post-attention gate receiving only an aggregate generally cannot distinguish its original item composition. Appendix~\ref{app:post_attention_path} provides a constructive example.

\subsection{Training Objective and Local Learning Signal}
\label{subsec:warrant_theoretical_properties}

Permission is learned jointly through the backbone's task loss. Evidence labels are not directly assigned as gate targets, and the gate-score bias initializes transmission strengths close to 1. OpenPath is a retrained control with gates fixed to 1 on the same path. When a path is added, OpenPath and the original Base are different models.

The local gradient reaching a permission logit is determined by the inner product between the current weighted value term and the loss gradient. Equation~\eqref{eq:warrant_gate_gradient} gives the derivative with respect to that gate logit while holding the other computations fixed.
\begin{equation}
\label{eq:warrant_gate_gradient}
\frac{\partial\mathcal{L}}{\partial\psi_j}
=(1-\lambda)\sigma'(\psi_j)
\left\langle\nabla_{h^W}\mathcal{L},c_j\right\rangle.
\end{equation}
In Eq.~\eqref{eq:warrant_gate_gradient}, a positive inner product means that increasing the contribution locally raises the loss, so gradient descent produces a signal to lower the gate logit. A negative inner product produces the opposite signal. Small attention weights and sigmoid saturation can weaken these learning signals.

Permission learns contribution transmission strengths through these loss-based signals. Appendix~\ref{app:theoretical_derivations} presents the local loss bound, SNR condition, manifold-local extension, entangled-contribution analysis, and bounded parameterization. Same-path performance comparisons and item-level analyses evaluate the effects of the learned control.

\subsection{Prediction-Path Construction}
\label{subsec:domain_adaptation}

Applying permission to each backbone requires connecting item-wise contributions to the prediction object they affect. We expose the item axis on an existing path or construct an attention-derived residual, using the same connection for OpenPath and Full. We refer to this implementation procedure as path localization. Access to the computation graph and prediction tensors allows the connection to be checked. Appendix~\ref{subsec:path_localization} provides the tracing protocol and tensor mappings.

The same item-wise operation is used in every task. The controlled term in Eq.~\eqref{eq:warrant_operator_map} is transmitted to the backbone's score or state interface.
\begin{equation}
\label{eq:warrant_operator_map}
\alpha_{ij}v_j \longmapsto \alpha_{ij}g(q_i,k_j)v_j.
\end{equation}
In Eq.~\eqref{eq:warrant_operator_map}, the query and item representations and the receiving prediction interface vary, while the definition of permission remains fixed.

\paragraph{CTDG link prediction.}
The CTDG task is continuous-time dynamic graph link prediction: predicting whether the current source--destination edge exists from a time-ordered interaction history. Because the prediction target is the current edge score rather than a token or node representation, the \method{} query represents an edge-level prediction request. An edge-query adapter constructs the current edge query, and \method{} gates weighted temporal-history terms from the source and destination sides. This interface exposes model-internal representations already used by the base temporal encoder to \method{}'s query--key inputs.

\paragraph{MTPP next-mark ranking.}
The MTPP task is marked temporal point process next-mark ranking, with Mark MRR as the primary metric over candidate marks. The same past event can provide support, a shortcut, or an ambiguous signal depending on the candidate mark. Scoring all candidates from a shared sequence state can weaken candidate-specific permission differences. \method{} forms a candidate query from the candidate-mark embedding and current sequence state, reads event-history values to form a candidate-specific warranted value, and supplies a residual to the candidate-mark logit. At this interface, \method{} controls whether a past marked-event value contributes to the current candidate-mark score.

\paragraph{RAG supporting evidence selection.}
The RAG task is retrieval-augmented supporting evidence selection. Evaluation concerns evidence selection rather than answer generation. Using HotpotQA's sentence-level supporting facts, we define a retrieved passage containing an annotated supporting fact as a supporting evidence passage. The primary metric, Support MRR, measures its rank. A distractor passage can receive attention through lexical overlap with the question without deserving a contribution to the supporting-evidence ranking score. \method{} forms warranted passage mass from the question query and retrieved passage representations and connects this mass directly to the support logit. It therefore gates the path through which weighted terms from supporting and distractor passages directly affect the ranking metric.

\paragraph{STPP next-location forecasting.}
The STPP task is spatio-temporal point process next-location forecasting. The model predicts the time and continuous location of the next event, with Location RMSE as the primary metric. Unlike CTDG and TKG settings that directly add biases to discrete candidate scores, STPP uses a continuous next-location dynamics state to determine the final metric. Strongly copying past coordinates can introduce local-cluster bias, and time likelihood and Location RMSE can move in different directions. \method{} therefore gates the path through which a prediction query reads historical values to form the next-event dynamics state, rather than acting as a coordinate-copy mechanism.

\paragraph{TKG tail prediction.}
The TKG task is temporal knowledge graph tail prediction. A query is a tuple of head, relation, and time; key--value items are historical temporal facts; and the primary metric is tail-entity ranking MRR. Historical facts can provide strong support, but a previously recurring tail need not be correct for the current query. The relevant bottleneck is the path from historical-fact values to tail scores or the copy path. For RE-NET and xERTE, warranted mass is added to historical tail-entity scores; for CyGNet, \method{} is connected to the path forming copy mass. This placement directly gates whether each historical fact's weighted tail contribution affects the current tail ranking.

\section{Experiments}
\label{sec:experiments}

The experiments address three questions: when ungated contributions impair prediction, whether learned permission selectively suppresses distractors while improving prediction, and how it compares with alternative controls. We first analyze CyGNet and HotpotQA, then evaluate the scope of the method through results across five task families and contribution interventions.

\subsection{Main Benchmark Settings}
\label{subsec:main_benchmark_settings}

All comparisons use the same reference implementation, seed set, and dataset--model grid, changing only the Base or \method{} variant. Table~\ref{tab:main_benchmark_settings} summarizes the training, \method{}, and model settings used in the reported benchmarks. Models and datasets follow the primary references in each field. CTDG uses the JODIE interaction datasets and TGAT, DyGFormer, and GraphMixer \cite{kumar2019jodie,xu2020tgat,yu2023dygformer,cong2023graphmixer}. MTPP uses the StackOverflow/Retweets settings in EasyTPP with SAHP, THP, and AttNHP \cite{xue2023easytpp,zhang2020self,zuo2020transformer,yang2022transformer}. RAG uses HotpotQA with FiD and an LED-style long-context reader \cite{yang2018hotpotqa,izacard2021leveraging,beltagy2020longformer}. STPP uses Earthquake and Gowalla with Transformer-STPP, NSTPP, and DeepSTPP \cite{chen2021neuralstpp,cho2011friendship,zhou2022deep}. TKG uses ICEWS/GDELT with RE-NET, xERTE, and CyGNet \cite{boschee2015icews,garcia2018learning,leetaru2013gdelt,jin2020renet,han2020xerte,zhu2021cygnet}.

\begin{center}
\begin{minipage}{\textwidth}
\centering
\captionof{table}{Main benchmark settings. Base and \method{} variants use the same reference implementation, seed set, and model grid.}
\label{tab:main_benchmark_settings}
\scriptsize
\resizebox{\textwidth}{!}{
\begin{tabular}{lll}
\toprule
Group & Setting & Value \\
\midrule
Experiment & seeds & \(7,17,37\) \\
\midrule
Training & epochs / max examples & 30; 65,536 \\
Training & optimizer & learning rate 0.001; weight decay 0.0 \\
Training & history / evaluation & history length 50; evaluation ratio 0.2 \\
\midrule
\method{} & context / auxiliary loss & no context schema or fields; auxiliary weight 0.0 \\
\method{} & gate initialization / leak & gate init 0.95; gate leak 0.05 \\
\method{} & domain loss weights & STPP gate loss 0.0; TKG gate loss 0.0 \\
\method{} & RAG exception & FiD/LED use gate leak 0.7 and support loss weight 1.0 \\
\midrule
Models & CTDG & Wikipedia, MOOC, LastFM with TGAT, DyGFormer, GraphMixer \\
Models & MTPP & StackOverflow, Retweets with SAHP, THP, AttNHP \\
Models & STPP & Earthquake, Gowalla with Transformer-STPP, NSTPP, DeepSTPP \\
Models & TKG & ICEWS14, ICEWS18, GDELT with RE-NET, xERTE, CyGNet \\
Models & RAG & HotpotQA with FiD, LED \\
\bottomrule
\end{tabular}
}
\end{minipage}
\end{center}

Base is the original backbone. OpenPath constructs the path on which permission is applied but fixes the gate to 1, whereas Full learns permission on the same path. Base--OpenPath measures the effect of path construction, and OpenPath--Full measures the additional effect of learned permission on that path. Each model is trained separately under matched conditions.

\subsection{When Direct Contributions Hurt}
\label{subsec:relevance_permission_separation}

We evaluate whether transmitting additional contributions to prediction is always beneficial and what difference permission makes on the same path. For all 32 dataset--model combinations in the main benchmark, we train OpenPath with the same data budget as Base and Full, 30 epochs, and seeds 7/17/37. Table~\ref{tab:cross_domain_path_permission} reports the complete results from these 96 additional runs.

{
\begin{center}
\begin{minipage}{\textwidth}
\centering
\captionof{table}{Main-budget decomposition of metric-facing path exposure and learned permission. All values are means over seeds 7, 17, and 37 under the same 30-epoch protocol as the main benchmark. OpenPath exposes the same path as Full with $g=1$. Deltas are oriented toward metric improvement; lower STPP RMSE is better.}
\label{tab:cross_domain_path_permission}
\scriptsize
\setlength{\tabcolsep}{3.2pt}
\renewcommand{\arraystretch}{1.08}
\resizebox{\textwidth}{!}{
\begin{tabular}{lllrrrrr}
\toprule
Domain & Dataset & Model & Base & OpenPath & Full & $\Delta_{path}$ & $\Delta_{perm}$ \\
\midrule
CTDG & LastFM & DyGFormer & .8482 & .9010 & .9023 & +.0528 & +.0013 \\
 & LastFM & GraphMixer & .8806 & .9142 & .9139 & +.0337 & -.0003 \\
 & LastFM & TGAT & .8598 & .8898 & .8897 & +.0300 & -.0002 \\
 & MOOC & DyGFormer & .9763 & .9768 & .9773 & +.0005 & +.0005 \\
 & MOOC & GraphMixer & .9770 & .9780 & .9774 & +.0010 & -.0006 \\
 & MOOC & TGAT & .9639 & .9716 & .9715 & +.0077 & -.0001 \\
 & Wikipedia & DyGFormer & .9440 & .9823 & .9836 & +.0383 & +.0014 \\
 & Wikipedia & GraphMixer & .9529 & .9834 & .9832 & +.0305 & -.0002 \\
 & Wikipedia & TGAT & .9437 & .9825 & .9824 & +.0389 & -.0002 \\
\midrule
MTPP & Retweets & AttNHP & .7574 & .7631 & .7653 & +.0057 & +.0022 \\
 & Retweets & SAHP & .7541 & .7577 & .7578 & +.0035 & +.0001 \\
 & Retweets & THP & .7584 & .7460 & .7497 & -.0124 & +.0037 \\
 & StackOverflow & AttNHP & .6051 & .6040 & .6041 & -.0011 & +.0001 \\
 & StackOverflow & SAHP & .6076 & .6063 & .6062 & -.0013 & -.0001 \\
 & StackOverflow & THP & .6066 & .6077 & .6076 & +.0011 & -.0001 \\
\midrule
RAG & HotpotQA & FiD & .6555 & .6565 & .6574 & +.0010 & +.0009 \\
 & HotpotQA & LED & .6669 & .6701 & .6670 & +.0032 & -.0031 \\
\midrule
STPP & Earthquake & DeepSTPP & 1.1383 & 1.0548 & 1.0502 & +.0835 & +.0046 \\
 & Earthquake & NSTPP & 1.1589 & 1.0774 & 1.1159 & +.0816 & -.0386 \\
 & Earthquake & Transformer-STPP & 1.1147 & 1.0754 & 1.0748 & +.0394 & +.0005 \\
 & Gowalla & DeepSTPP & .0481 & .0484 & .0484 & -.0002 & +.0000 \\
 & Gowalla & NSTPP & .0486 & .0475 & .0474 & +.0011 & +.0001 \\
 & Gowalla & Transformer-STPP & .0483 & .0460 & .0460 & +.0023 & +.0000 \\
\midrule
TKG & GDELT & CyGNet & .1478 & .1453 & .1478 & -.0025 & +.0025 \\
 & GDELT & RE-NET & .1280 & .1363 & .1395 & +.0083 & +.0032 \\
 & GDELT & xERTE & .1277 & .1389 & .1429 & +.0112 & +.0039 \\
 & ICEWS14 & CyGNet & .2195 & .2102 & .2196 & -.0093 & +.0094 \\
 & ICEWS14 & RE-NET & .2076 & .2141 & .2127 & +.0065 & -.0014 \\
 & ICEWS14 & xERTE & .2096 & .2183 & .2176 & +.0087 & -.0007 \\
 & ICEWS18 & CyGNet & .1536 & .1438 & .1501 & -.0098 & +.0063 \\
 & ICEWS18 & RE-NET & .1423 & .1520 & .1550 & +.0098 & +.0030 \\
 & ICEWS18 & xERTE & .1515 & .1570 & .1580 & +.0055 & +.0010 \\
\bottomrule
\end{tabular}
}
\end{minipage}
\end{center}
}

For CyGNet, adding an ungated path reduces MRR on all three datasets: from .1478 for Base to .1453 for OpenPath on GDELT, from .2195 to .2102 on ICEWS14, and from .1536 to .1438 on ICEWS18. Full obtains .1478, .2196, and .1501 on the same paths, recovering .0025, .0094, and .0063 MRR over OpenPath, respectively. Recovery on ICEWS18 is partial and does not surpass Base. Since CyGNet already has a historical copy path, OpenPath here denotes the additional contribution interface constructed in this experiment, rather than the presence or absence of its original copy mechanism.

These results show that learning contribution transmission strength changes prediction even with the same connection. Across TKG as a whole, OpenPath is beneficial on average. Over 27 matched-seed comparisons, the mean path effect is +.0032 MRR and the additional permission effect is +.0030 MRR; the latter is positive in 21 comparisons. The three CyGNet settings provide repeated instances of ungated-path degradation and recovery through permission.

Retweets/THP also exhibits degradation from path addition and improvement through permission, and permission yields additional gains for every Retweets backbone. On HotpotQA/FiD, path construction and permission increase Support MRR by +.0010 and +.0009, respectively. The next section uses labeled evidence to analyze which items receive lower transmission strength.

\subsection{Learning to Suppress Distractors}
\label{subsec:pretrained_transformer_permission}

We jointly measure learned permission and changes in evidence ranking on HotpotQA supporting sentence selection. Candidate sets containing annotated support and distractors allow us to analyze item-level gates, ranking, and unsupported attribution. Appendix~\ref{app:hotpotqa_roberta} provides detailed settings and complete diagnostics.

We fine-tune \texttt{roberta-base}~\cite{liu2019roberta} for HotpotQA supporting sentence selection. Each input contains a question, an instruction, up to eight candidate evidence sentences, and candidate markers. Candidate sets include annotated supporting sentences, hard distractors, and random distractors; the primary metric is Support MRR. \texttt{openpath\_nogate} opens the same metric-facing candidate-marker value path with \(g=1\), whereas \texttt{full\_warrant} replaces the item-wise term \(\alpha_{ij}v_j\) on that path with \(\alpha_{ij}g_{ij}v_j\). Appendix~\ref{app:hotpotqa_roberta} gives the full setup and permission diagnostics.

\paragraph{Learned permission and prediction changes.}
Across five matched seeds on HotpotQA/RoBERTa, we measure attention mass and gates by item type and compare Support MRR and Unsupported@GoldCount on the same path.

\begin{center}
\begin{minipage}{\textwidth}
\centering
\small
\captionof{table}{Learned distractor suppression and prediction changes on HotpotQA/RoBERTa (5 matched seeds). The upper block reports means by item type; the lower block reports metric means and standard deviations across seeds. Full assigns lower gates to distractors than to gold items and obtains higher MRR than OpenPath in four of five seeds. Higher Support MRR and lower Unsupported@GoldCount are better.}
\label{tab:harmful_path_recovery}
\vspace{2mm}
\begin{tabular}{lccc}
\toprule
Signal & Gold support & Hard distractor & Random distractor \\
\midrule
OpenPath attention mass & .0129 & .0136 & .0194 \\
Full permission gate & .8882 & .5054 & .4665 \\
\bottomrule
\end{tabular}
\vspace{2mm}

\begin{tabular}{lcc}
\toprule
Variant & Support MRR & Unsupported@GoldCount \\
\midrule
OpenPath & .9111$\pm$.0084 & .2271$\pm$.0117 \\
Full & \textbf{.9141$\pm$.0045} & \textbf{.2215$\pm$.0072} \\
\bottomrule
\end{tabular}
\end{minipage}
\end{center}

In Table~\ref{tab:harmful_path_recovery}, OpenPath assigns attention masses of .0136 and .0194 to hard and random distractors, respectively, exceeding the gold mass of .0129. Full maintains a gold gate of .8882 while reducing hard and random gates to .5054 and .4665. In the same experiment, mean Support MRR increases by +.0030 and Unsupported@GoldCount decreases by .0056; Full outperforms OpenPath in four of five seeds.

Support MRR changes from .9111$\pm$.0084 to .9141$\pm$.0045, while Unsupported@GoldCount changes from .2271$\pm$.0117 to .2215$\pm$.0072. The latter is the fraction of unsupported sentences among as many selected sentences as there are gold supports; lower is better. Lower distractor permissions coincide with higher Support MRR and lower unsupported selection.

\paragraph{Encoder and readout controls.}
\begin{center}
\begin{minipage}{\textwidth}
\centering
\captionof{table}{HotpotQA/RoBERTa candidate-marker controls. All variants use the same candidate-marker inputs, split, metrics, and three seeds. Lower Unsupported@GoldCount is better.}
\label{tab:hotpotqa_roberta_controls}
\scriptsize
\setlength{\tabcolsep}{3.0pt}
\renewcommand{\arraystretch}{1.12}
\resizebox{\textwidth}{!}{
\begin{tabular}{lrrrrr}
\toprule
Variant & MRR \(\uparrow\) & R@1 \(\uparrow\) & Evidence F1 \(\uparrow\) & AUPRC \(\uparrow\) & Unsupported@GoldCount \(\downarrow\) \\
\midrule
\texttt{base} & \(0.9111 \pm 0.0045\) & \(0.8371 \pm 0.0073\) & \(0.7751 \pm 0.0049\) & \(0.8110 \pm 0.0065\) & \(0.2249 \pm 0.0049\) \\
\texttt{param\_mlp} & \(0.9036 \pm 0.0057\) & \(0.8245 \pm 0.0106\) & \(0.7688 \pm 0.0033\) & \(0.8103 \pm 0.0082\) & \(0.2312 \pm 0.0033\) \\
\texttt{post\_attention\_glu} & \(0.9051 \pm 0.0039\) & \(0.8295 \pm 0.0098\) & \(0.7629 \pm 0.0084\) & \(0.8034 \pm 0.0116\) & \(0.2371 \pm 0.0084\) \\
\texttt{attention\_readout} & \(0.9125 \pm 0.0042\) & \(0.8405 \pm 0.0072\) & \(0.7730 \pm 0.0076\) & \(\mathbf{0.8129} \pm 0.0040\) & \(0.2270 \pm 0.0076\) \\
\texttt{query\_only\_gate} & \(0.9078 \pm 0.0021\) & \(0.8312 \pm 0.0048\) & \(0.7726 \pm 0.0057\) & \(0.8104 \pm 0.0053\) & \(0.2274 \pm 0.0057\) \\
\texttt{shuffled\_warrant} & \(0.9046 \pm 0.0102\) & \(0.8270 \pm 0.0166\) & \(0.7671 \pm 0.0160\) & \(0.7979 \pm 0.0086\) & \(0.2329 \pm 0.0160\) \\
\texttt{openpath\_nogate} & \(0.9129 \pm 0.0092\) & \(0.8405 \pm 0.0165\) & \(0.7755 \pm 0.0110\) & \(0.8102 \pm 0.0127\) & \(0.2245 \pm 0.0110\) \\
\texttt{full\_warrant} & \(\mathbf{0.9134} \pm 0.0010\) & \(\mathbf{0.8414} \pm 0.0052\) & \(\mathbf{0.7772} \pm 0.0058\) & \(0.8042 \pm 0.0027\) & \(\mathbf{0.2228} \pm 0.0058\) \\
\bottomrule
\end{tabular}
}
\end{minipage}
\end{center}

The separate three-seed control in Table~\ref{tab:hotpotqa_roberta_controls} examines added parameters, post-attention gating, and query--item pairing. \texttt{full\_warrant} exceeds \texttt{param\_mlp} by +0.0098 MRR, \texttt{post\_attention\_glu} by +0.0083 MRR, and \texttt{shuffled\_warrant}, which breaks query--item pairing, by +0.0088 MRR. The strong \texttt{openpath\_nogate} result also shows that opening the metric-facing candidate-marker value path explains an important part of the gains in this HotpotQA/RoBERTa setting. \texttt{full\_warrant} adds a small learned-permission refinement, obtaining the best MRR, R@1, Evidence F1, and Unsupported@GoldCount. Its AUPRC is below \texttt{base}. We therefore interpret this control in terms of improved top-end ranking and unsupported evidence attribution rather than uniform improvement across metrics.

\paragraph{Unsupported evidence attribution.}
Unsupported@GoldCount measures the fraction of unsupported candidates selected when the number of selected evidence sentences equals the gold support count. It captures evidence-attribution failures in which unsupported sentences are selected as supporting evidence within a controlled candidate set; the evaluation concerns candidate evidence selection. \texttt{full\_warrant} reduces Unsupported@GoldCount from 0.2249 to 0.2228, an approximately 0.94\% reduction in unsupported-attribution error. Together with increases in MRR, R@1, and Evidence F1, this indicates that \method{} reduces unsupported evidence selection in this setting.

\subsection{Comparison with Alternative Controls}
\label{subsec:permission_function_controls}

To analyze the control freedoms provided by permission, we compare four mechanisms in representative main-budget settings from five domains. Scalar applies one query-specific value to all items, whereas item-only computes per-item gates without the current query. Normalized renormalizes \(\alpha_{ij}g_{ij}\) across items, fixing total mass and changing only relative allocation. The attention adapter reweights attention logits using the same inputs and similar additional capacity. All variants use the same metric-facing path, 30 epochs, and seeds 7/17/37.

\begin{center}
\begin{minipage}{\textwidth}
\centering
\small
\setlength{\tabcolsep}{3.5pt}
\captionof{table}{Main-budget controls decomposing permission functions on the same metric-facing path. Full uses the existing three-seed main decomposition; the other controls are trained for the same 30 epochs with seeds 7/17/37. Lower is better for STPP; higher is better for all other tasks.}
\label{tab:permission_function_controls}
\vspace{2mm}
\begin{tabular}{llccccc}
\toprule
Domain & Dataset / model & Scalar & Item-only & Attention adapter & Normalized & Full \\
\midrule
CTDG & LastFM / DyGFormer & .9026$\pm$.0018 & .9022$\pm$.0022 & .9032$\pm$.0006 & \textbf{.9061$\pm$.0008} & .9023 \\
MTPP & Retweets / THP & \textbf{.7574$\pm$.0153} & .7503$\pm$.0127 & .7543$\pm$.0081 & .7520$\pm$.0126 & .7497 \\
RAG & HotpotQA / FiD & .6588$\pm$.0008 & .6583$\pm$.0015 & .6589$\pm$.0015 & \textbf{.6593$\pm$.0007} & .6574 \\
STPP$\downarrow$ & Earthquake / DeepSTPP & 1.0533$\pm$.0942 & 1.0588$\pm$.0780 & 1.0564$\pm$.0717 & 1.0623$\pm$.0806 & \textbf{1.0502} \\
TKG & GDELT / xERTE & .1425$\pm$.0008 & .1424$\pm$.0007 & .1426$\pm$.0003 & \textbf{.1434$\pm$.0006} & .1429 \\
\bottomrule
\end{tabular}
\end{minipage}
\end{center}

In Table~\ref{tab:permission_function_controls}, the normalized variant performs best on CTDG, RAG, and TKG, showing that relative allocation alone can be effective. Scalar performs best on MTPP, where total-mass control alone is competitive. On STPP, Full, which jointly controls relative allocation and total mass, achieves the lowest RMSE. The attention adapter is also competitive on CTDG, RAG, and TKG. Effective control freedoms vary by task even on the same metric-facing path; Full's unnormalized query--item gate provides both relative allocation and total-mass control.

\subsubsection{Comparison with G1/G2 Gated Attention}
\label{subsec:qiu_gated_attention_comparison}

G1 and G2 from Qiu et al.~\cite{qiu2025gated} provide the closest gated-attention placements to Warrant. Following the original definitions, we implement head-specific element-wise sigmoid gates in every RoBERTa self-attention layer. G2 computes a gate from each key/value token's hidden state and multiplies the value projection before the attention-weighted sum. G1 first computes the SDPA output and then multiplies the aggregate by a gate obtained from the current query token's hidden state. Full Warrant computes a gate for the current query--item pair at the final candidate-marker layer and applies it to the localized metric-facing weighted-value term before aggregation.

\begin{center}
\begin{minipage}{\textwidth}
\centering
\small
\captionof{table}{Qiu et al.'s G1/G2 versus Full Warrant on HotpotQA/RoBERTa (5 seeds; mean$\pm$std). G2 gates each layer's value projection using the key/value-token hidden state, whereas G1 gates each layer's SDPA output using the query-token hidden state. Qiu gates use the default optimizer learning rate; Warrant uses its existing gate learning-rate setting. All methods use the same split and train for 2 epochs.}
\label{tab:qiu_gated_attention_comparison}
\vspace{2mm}
\resizebox{\textwidth}{!}{%
\begin{tabular}{lrrrrrr}
\toprule
Variant & Params & Support MRR & R@1 & Evidence F1 & Unsupported $\downarrow$ & AUPRC \\
\midrule
Base & 124.65M & $.9083\pm.0087$ & $.8314\pm.0161$ & $.7719\pm.0076$ & $.2281\pm.0076$ & $.8043\pm.0202$ \\
OpenPath & 129.97M & $.9111\pm.0084$ & $.8390\pm.0145$ & $.7729\pm.0117$ & $.2271\pm.0117$ & $.8125\pm.0119$ \\
Qiu G2 & 131.74M & $.9094\pm.0058$ & $.8344\pm.0120$ & $.7747\pm.0024$ & $.2253\pm.0024$ & $.8126\pm.0069$ \\
Qiu G1 & 131.74M & $.9090\pm.0041$ & $.8349\pm.0083$ & $.7704\pm.0145$ & $.2296\pm.0145$ & $.8048\pm.0149$ \\
Full Warrant & 129.97M & $\mathbf{.9122\pm.0070}$ & $\mathbf{.8395\pm.0141}$ & $\mathbf{.7770\pm.0090}$ & $\mathbf{.2230\pm.0090}$ & $\mathbf{.8138\pm.0098}$ \\
\bottomrule
\end{tabular}%
}
\end{minipage}
\end{center}

In Table~\ref{tab:qiu_gated_attention_comparison}, Full Warrant obtains mean Support MRR .9122, R@1 .8395, Evidence F1 .7770, and AUPRC .8138, with Unsupported@GoldCount .2230. Qiu G2 and G1 obtain Support MRR .9094 and .9090, respectively, while Full uses fewer parameters than either method. The seed-wise Full--G2 MRR difference averages $+.0028\pm.0119$ and is positive in three of five seeds; the Full--G1 difference is $+.0033\pm.0046$ and is positive in four seeds. G2 preserves item identity, whereas G1 controls the aggregate using the query. By using both query and item to control metric-facing terms, Full obtains strong mean ranking and evidence-attribution results in this comparison. Full has the highest mean Support MRR, with variable seed-wise differences. This table reports a five-seed comparison conducted separately from the recovery experiment in Section~\ref{subsec:pretrained_transformer_permission}.

\paragraph{Frozen role control.}
Starting from OpenPath checkpoints on RAG and TKG, we freeze the encoder and readout and train only the gate or a capacity-matched attention adapter. This measures the changes produced by the two control modules on fixed representations and readouts.

\begin{center}
\begin{minipage}{\textwidth}
\centering
\small
\captionof{table}{Control modules on fixed OpenPath representations. The encoder and readout of the same checkpoint are frozen, and only the gate or a capacity-matched attention adapter is trained.}
\label{tab:frozen_role_controls}
\vspace{2mm}
\begin{tabular}{llccc}
\toprule
Domain & Dataset / model & Frozen OpenPath & + Gate only & + Attention adapter only \\
\midrule
RAG & HotpotQA / FiD & .6565$\pm$.0005 & .6570$\pm$.0004 & \textbf{.6575$\pm$.0006} \\
TKG & GDELT / xERTE & \textbf{.1426$\pm$.0008} & .1424$\pm$.0010 & .1420$\pm$.0011 \\
\bottomrule
\end{tabular}
\end{minipage}
\end{center}

On RAG, both modules slightly improve frozen OpenPath, with a larger gain from the attention adapter. On TKG, both changes from the frozen starting point are negative. The modules control normalized access and post-softmax weighted-value scaling, respectively, and their effects on fixed representations also vary by setting. The learned-gate analysis in Section~\ref{subsec:pretrained_transformer_permission} and the interventions in Section~\ref{subsec:evidence_mass_diagnostics} measure contribution control from complementary perspectives.

\subsection{Cross-Domain Evaluation}
\label{subsec:cross_domain_performance}

The main benchmark covers five task families: CTDG, MTPP, RAG, STPP, and TKG. We conduct 32 paired Base-versus-\method{} comparisons, reporting means and standard deviations across three seeds for each comparison, for a total of 192 completed runs. Table~\ref{tab:benchmark} reports the paired results. Each row compares Base and \method{} on the same domain, dataset, and model, with improvement calculated according to metric direction. The \(\pm\) values in Table~\ref{tab:benchmark} are seed-wise standard deviations. Results are summarized by improvement direction and the practical tiers defined in the table.

{
\newcommand{\gain}[1]{\textcolor{green!45!black}{\(\mathbf{+#1}\)}}
\newcommand{\loss}[1]{\textcolor{red!70!black}{\(\mathbf{-#1}\)}}
\newcommand{\domainrow}[2]{\rowcolor{blue!8}\multicolumn{8}{l}{\textbf{#1}\hfill\textit{#2}}\\}
\begin{center}
\begin{minipage}{\textwidth}
\centering
\captionof{table}{Complete main benchmark results by domain. Each row compares Base and \method{} on the same dataset and model using means and standard deviations over three seeds. Paired \(\Delta\) is the mean difference oriented toward improvement. The \(\pm\) values are seed-wise standard deviations, not row-level 95\% confidence intervals or paired \(t\)-tests. Tiers indicate practical interpretations.}
\label{tab:benchmark}
\scriptsize
\setlength{\tabcolsep}{4pt}
\renewcommand{\arraystretch}{1.12}
\resizebox{\textwidth}{!}{
\begin{tabular}{lllrrrrl}
\toprule
Dataset & Model & Metric & Base & \method{} & Paired \(\Delta\) & Relative & Tier \\
\midrule
\domainrow{CTDG}{continuous-time dynamic graph link prediction, AUC \(\uparrow\)}
LastFM & DyGFormer & AUC & \(0.8482 \pm 0.0045\) & \(0.9023 \pm 0.0022\) & \gain{0.0541} & \gain{6.38\%} & substantial \\
LastFM & GraphMixer & AUC & \(0.8806 \pm 0.0031\) & \(0.9139 \pm 0.0009\) & \gain{0.0334} & \gain{3.79\%} & substantial \\
LastFM & TGAT & AUC & \(0.8598 \pm 0.0044\) & \(0.8897 \pm 0.0024\) & \gain{0.0298} & \gain{3.47\%} & substantial \\
MOOC & TGAT & AUC & \(0.9639 \pm 0.0019\) & \(0.9715 \pm 0.0007\) & \gain{0.0076} & \gain{0.79\%} & positive uncertain \\
MOOC & DyGFormer & AUC & \(0.9763 \pm 0.0003\) & \(0.9773 \pm 0.0016\) & \gain{0.0010} & \gain{0.11\%} & tie/negligible \\
MOOC & GraphMixer & AUC & \(0.9770 \pm 0.0000\) & \(0.9774 \pm 0.0015\) & \gain{0.0004} & \gain{0.04\%} & tie/negligible \\
Wikipedia & DyGFormer & AUC & \(0.9440 \pm 0.0015\) & \(0.9836 \pm 0.0005\) & \gain{0.0396} & \gain{4.20\%} & substantial \\
Wikipedia & TGAT & AUC & \(0.9437 \pm 0.0027\) & \(0.9824 \pm 0.0010\) & \gain{0.0387} & \gain{4.10\%} & substantial \\
Wikipedia & GraphMixer & AUC & \(0.9529 \pm 0.0040\) & \(0.9832 \pm 0.0003\) & \gain{0.0303} & \gain{3.18\%} & substantial \\
\midrule
\domainrow{MTPP}{marked temporal point process next-mark ranking, Mark MRR \(\uparrow\)}
Retweets & AttNHP & Mark MRR & \(0.7574 \pm 0.0130\) & \(0.7653 \pm 0.0049\) & \gain{0.0079} & \gain{1.06\%} & positive uncertain \\
Retweets & SAHP & Mark MRR & \(0.7541 \pm 0.0092\) & \(0.7578 \pm 0.0106\) & \gain{0.0037} & \gain{0.49\%} & positive uncertain \\
Retweets & THP & Mark MRR & \(0.7584 \pm 0.0029\) & \(0.7497 \pm 0.0115\) & \loss{0.0086} & \loss{1.14\%} & drop \\
StackOverflow & THP & Mark MRR & \(0.6066 \pm 0.0017\) & \(0.6076 \pm 0.0013\) & \gain{0.0009} & \gain{0.15\%} & tie/negligible \\
StackOverflow & AttNHP & Mark MRR & \(0.6051 \pm 0.0020\) & \(0.6041 \pm 0.0045\) & \loss{0.0010} & \loss{0.17\%} & drop \\
StackOverflow & SAHP & Mark MRR & \(0.6076 \pm 0.0017\) & \(0.6062 \pm 0.0017\) & \loss{0.0013} & \loss{0.22\%} & drop \\
\midrule
\domainrow{RAG}{retrieval-augmented supporting evidence selection, Support MRR \(\uparrow\)}
HotpotQA & FiD & Support MRR & \(0.6555 \pm 0.0017\) & \(0.6574 \pm 0.0030\) & \gain{0.0019} & \gain{0.29\%} & positive uncertain \\
HotpotQA & LED & Support MRR & \(0.6669 \pm 0.0018\) & \(0.6670 \pm 0.0002\) & \gain{0.0002} & \gain{0.02\%} & tie/negligible \\
\midrule
\domainrow{STPP}{spatio-temporal point process next-location forecasting, Location RMSE \(\downarrow\)}
Earthquake & DeepSTPP & Location RMSE & \(1.1383 \pm 0.0288\) & \(1.0502 \pm 0.0926\) & \gain{0.0881} & \gain{7.72\%} & substantial \\
Earthquake & NSTPP & Location RMSE & \(1.1589 \pm 0.0612\) & \(1.1159 \pm 0.1006\) & \gain{0.0430} & \gain{3.84\%} & substantial \\
Earthquake & Transformer-STPP & Location RMSE & \(1.1147 \pm 0.0893\) & \(1.0748 \pm 0.0948\) & \gain{0.0399} & \gain{3.56\%} & substantial \\
Gowalla & Transformer-STPP & Location RMSE & \(0.0483 \pm 0.0047\) & \(0.0460 \pm 0.0024\) & \gain{0.0023} & \gain{4.38\%} & positive uncertain \\
Gowalla & NSTPP & Location RMSE & \(0.0486 \pm 0.0033\) & \(0.0474 \pm 0.0017\) & \gain{0.0012} & \gain{2.26\%} & tie/negligible \\
Gowalla & DeepSTPP & Location RMSE & \(0.0481 \pm 0.0012\) & \(0.0484 \pm 0.0046\) & \loss{0.0002} & \loss{0.42\%} & drop \\
\midrule
\domainrow{TKG}{temporal knowledge graph tail prediction, MRR \(\uparrow\)}
GDELT & xERTE & MRR & \(0.1277 \pm 0.0022\) & \(0.1429 \pm 0.0008\) & \gain{0.0151} & \gain{11.85\%} & substantial \\
GDELT & RE-NET & MRR & \(0.1280 \pm 0.0011\) & \(0.1395 \pm 0.0010\) & \gain{0.0115} & \gain{9.01\%} & marginal \\
GDELT & CyGNet & MRR & \(0.1478 \pm 0.0006\) & \(0.1478 \pm 0.0012\) & \gain{0.0000} & \gain{0.02\%} & tie/negligible \\
ICEWS14 & xERTE & MRR & \(0.2096 \pm 0.0024\) & \(0.2176 \pm 0.0032\) & \gain{0.0080} & \gain{3.81\%} & positive uncertain \\
ICEWS14 & RE-NET & MRR & \(0.2076 \pm 0.0008\) & \(0.2127 \pm 0.0046\) & \gain{0.0052} & \gain{2.50\%} & positive uncertain \\
ICEWS14 & CyGNet & MRR & \(0.2195 \pm 0.0039\) & \(0.2196 \pm 0.0031\) & \gain{0.0000} & \gain{0.02\%} & tie/negligible \\
ICEWS18 & RE-NET & MRR & \(0.1423 \pm 0.0002\) & \(0.1550 \pm 0.0025\) & \gain{0.0127} & \gain{8.95\%} & positive uncertain \\
ICEWS18 & xERTE & MRR & \(0.1515 \pm 0.0017\) & \(0.1580 \pm 0.0022\) & \gain{0.0065} & \gain{4.31\%} & tie/negligible \\
ICEWS18 & CyGNet & MRR & \(0.1536 \pm 0.0007\) & \(0.1501 \pm 0.0054\) & \loss{0.0035} & \loss{2.24\%} & drop \\
\bottomrule
\end{tabular}
}
\end{minipage}
\end{center}
}

Overall performance relative to Base reflects the joint application of path construction and permission. \method{} improves the primary metric in 27 of 32 comparisons, with gains spanning multiple domains and architectures. An exact two-sided sign test gives \(p=0.000113\), assuming independence across comparisons. Practical tiering assigns 10 comparisons to substantial, one to marginal, eight to positive but uncertain, eight to tie/negligible, and five to drop. AUC increases in 9/9 CTDG combinations, and MRR increases in 8/9 TKG combinations. Location RMSE decreases in five of six STPP combinations, and Support MRR increases for both HotpotQA readers in RAG. MTPP improves on Retweets but is mixed on StackOverflow.

Mean Base--Full improvements by domain are +.0261 AUC for CTDG, +.0002 Mark MRR for MTPP, +.0010 Support MRR for RAG, a .0290 reduction in Location RMSE for STPP, and +.0062 MRR for TKG.

In Table~\ref{tab:cross_domain_path_permission}, the 27 matched-seed CTDG comparisons have a mean path effect of +.0259 AUC and a permission effect of +.0002. Permission increments are positive for DyGFormer on LastFM, MOOC, and Wikipedia but small for GraphMixer and TGAT. STPP is also largely path-driven, while StackOverflow has nearly zero permission increments. These results distinguish overall Base--Full gains from gains attributable to permission alone. Appendix~\ref{subsec:path_localization_adaptation} analyzes path placement and the edge-query adapter.

Section~\ref{subsec:failure_regimes} analyzes the drop rows directly through their Base--OpenPath--Full changes.

\subsection{Contribution-Level Diagnostics}
\label{subsec:evidence_mass_diagnostics}

\subsubsection{Attention Relevance and Prediction Utility}
\label{subsec:marginal_value_audit}

Attention represents query--item compatibility, whereas whether a weighted-value term supports the correct prediction depends on its value direction and interactions with other items. To measure this distinction directly, we analyze frozen Base checkpoints for LastFM/DyGFormer, Retweets/AttNHP, HotpotQA/FiD, Earthquake/DeepSTPP, and GDELT/xERTE. Each checkpoint uses at most 10,000 training examples and 10 epochs; we evaluate 128 examples per seed for seeds 7/17/37. For each example, we select the five highest-attention items and evaluate all 32 subsets while keeping the remaining valid items fixed. Removing an item zeros its post-softmax weighted-value term without renormalizing attention, blocking only that contribution. We compute each item's marginal prediction utility as its exact Shapley value across all subset contexts. Utility is negative binary cross-entropy for CTDG, negative cross-entropy for MTPP and TKG, gold-passage log probability for RAG, and negative location error for STPP.

\begin{center}
\begin{minipage}{\textwidth}
\centering
\small
\captionof{table}{Marginal prediction utility of top-attended contributions in representative frozen Base checkpoints (3 seeds; mean$\pm$std). Each seed evaluates 128 examples. Attn.--utility $\rho$ is the Spearman correlation between attention and exact Shapley utility. Harmful top is the fraction of examples in which the highest-attention item has negative Shapley utility. Gain denotes metric increases for CTDG, MTPP, RAG, and TKG, and the Location RMSE reduction for STPP.}
\label{tab:marginal_value_audit}
\vspace{2mm}
\resizebox{\textwidth}{!}{%
\begin{tabular}{llrrrr}
\toprule
Domain & Dataset / model & Attn.--utility $\rho$ & Harmful top & Harmful items & Gain $\uparrow$ \\
\midrule
CTDG & LastFM / DyGFormer & $.094\pm.071$ & $43.5\pm7.3\%$ & $42.7\pm7.2\%$ & $+.0351\pm.0099$ \\
MTPP & Retweets / AttNHP & $.019\pm.019$ & $45.1\pm10.7\%$ & $44.7\pm9.5\%$ & $+.0130\pm.0135$ \\
RAG  & HotpotQA / FiD & $-.083\pm.066$ & $54.2\pm5.1\%$ & $49.9\pm2.5\%$ & $+.0425\pm.0112$ \\
STPP & Earthquake / DeepSTPP & $.083\pm.254$ & $50.0\pm27.8\%$ & $42.2\pm13.9\%$ & $+.0092\pm.0068$ \\
TKG  & GDELT / xERTE & $-.010\pm.046$ & $54.4\pm3.0\%$ & $53.6\pm1.0\%$ & $+.0001\pm.0002$ \\
\bottomrule
\end{tabular}%
}
\end{minipage}
\end{center}

In Table~\ref{tab:marginal_value_audit}, mean seed-wise attention--utility rank correlations range from $-.083$ to $.094$ across all five settings. The highest-attention item's contribution has negative Shapley utility in 43.5\% of CTDG examples, 45.1\% of MTPP examples, 54.2\% of RAG examples, 50.0\% of STPP examples, and 54.4\% of TKG examples. Across all top-five items, an average of 42.2--53.6\% of contributions are negative. STPP correlations and harmful fractions vary substantially across seeds, but all three seeds contain harmful contributions and have positive best-subset gains. High attention does not imply high prediction utility: in all five representative settings, some attention-accessed items move the prediction object against the current target.

Selecting top-five subsets using ground-truth utility improves mean CTDG AUC by .0351, MTPP Mark MRR by .0130, RAG Support MRR by .0425, and TKG MRR by .0001, while reducing STPP Location RMSE by .0092 relative to retaining all contributions. Gains in Table~\ref{tab:marginal_value_audit} use a positive sign for improvement in every domain. They measure the remaining scope for selective control within a fixed checkpoint and its selected top-five items. These gains come from oracle selection using ground truth and are distinct from learned permission performance.

\subsubsection{Evidence-Mass Intervention}

In retrieval, a passage with high lexical overlap with the question may receive attention without providing supporting evidence. In TKG, a historical fact may be temporally and structurally related to a query without supporting the true tail in the current ranking. In CTDG, temporal history may be relevant to the current edge while counterpart evidence and surrounding context contribute with different strengths. \method{} uses permission to regulate how these weighted-value terms contribute to the current prediction object.

We examine these effects through a limited neural-dissection experiment separate from the main benchmark. For one dataset--model pair per domain, we record gate behavior, warranted mass, and contribution paths with seed 7, 10 epochs, and at most 8192 examples. This diagnostic tests whether attention mass and warranted mass separate and whether the localized contribution path exhibits counterfactual dependency on labeled evidence.

Figure~\ref{fig:mass_separation} summarizes the observations. In RAG, warranted support mass increases and distractor mass decreases while the support attention share remains nearly unchanged. This change in effective mass is not captured by interpreting the attention map alone. In the TKG panel, the localized interface forms a historical-tail contribution path, separating tail ranking from Base even when gates are near identity. Gate magnitude is therefore interpreted together with interface placement and actual mass changes.

\begin{center}
\includegraphics[width=\textwidth]{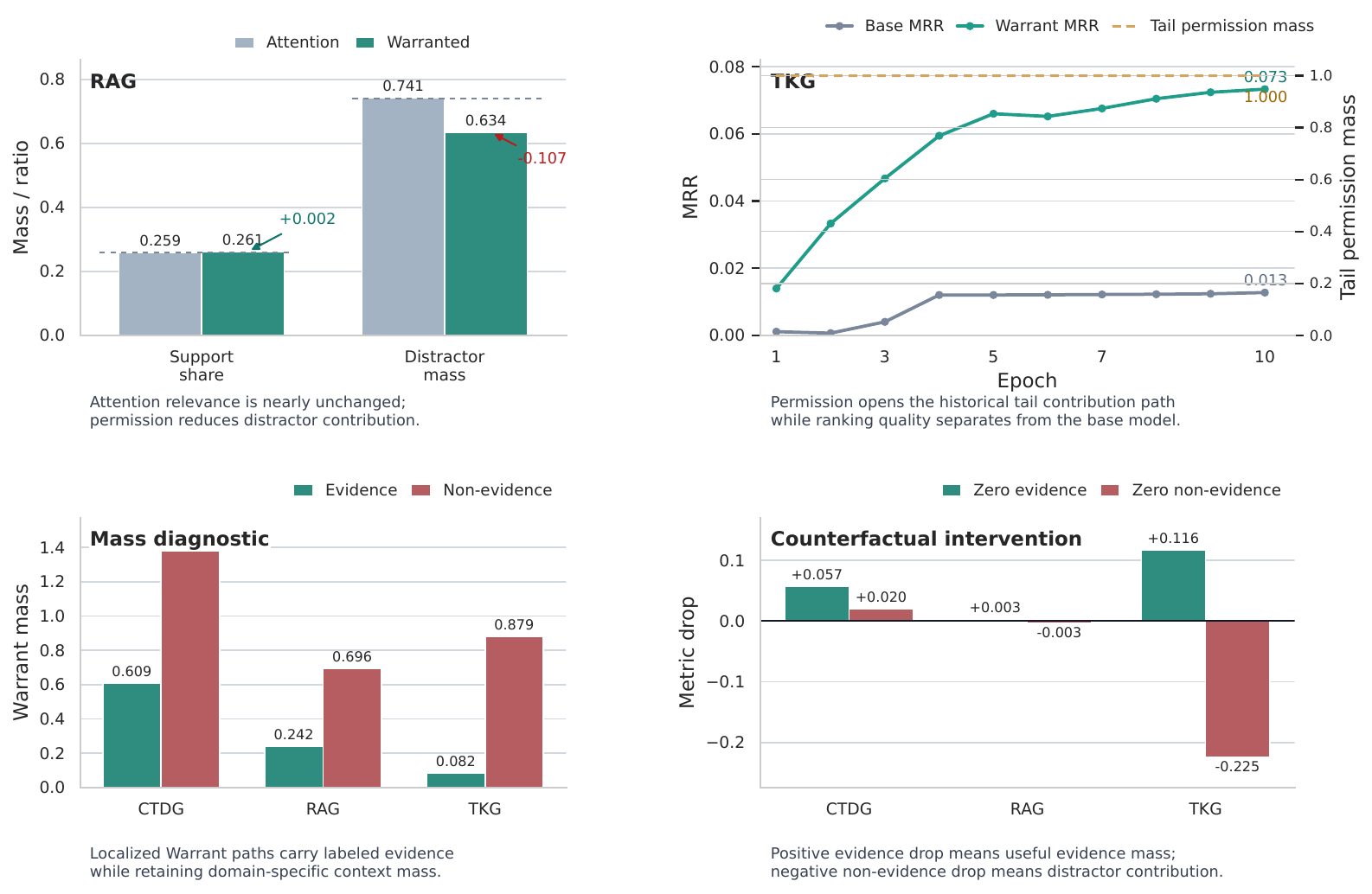}
\captionof{figure}{Mass separation and counterfactual mass diagnostics. The upper panels show RAG relevance--permission separation and TKG path exposure; the lower panels show zeroing interventions for CTDG, RAG, and TKG.}
\label{fig:mass_separation}
\end{center}

Table~\ref{tab:mass_diagnostic} reports the corresponding intervention values.

\begin{center}
\begin{minipage}{\textwidth}
\centering
\captionof{table}{Mass diagnostics using evaluation-time counterfactual interventions. A positive zero-evidence drop indicates useful evidence mass; a negative zero-non-evidence drop indicates that non-evidence contributions impair prediction under the oracle counterfactual.}
\label{tab:mass_diagnostic}
\resizebox{\textwidth}{!}{
\begin{tabular}{llrrrrr}
\toprule
Domain & Metric & Primary & Evidence mass & Non-evidence mass & Drop: zero evidence & Drop: zero non-evidence \\
\midrule
CTDG & AUC & \(0.8916 \pm 0.0073\) & \(0.6086 \pm 0.0234\) & \(1.3791 \pm 0.0236\) & \(+0.0575 \pm 0.0068\) & \(+0.0203 \pm 0.0012\) \\
RAG & Support MRR & \(0.5527 \pm 0.0113\) & \(0.2419 \pm 0.0008\) & \(0.6960 \pm 0.0082\) & \(+0.0030 \pm 0.0015\) & \(-0.0034 \pm 0.0008\) \\
TKG & MRR & \(0.1173 \pm 0.0022\) & \(0.0815 \pm 0.0029\) & \(0.8787 \pm 0.0308\) & \(+0.1162 \pm 0.0022\) & \(-0.2245 \pm 0.0027\) \\
\bottomrule
\end{tabular}
}
\end{minipage}
\end{center}

In RAG, removing supporting-passage mass reduces Support MRR by 0.0030, whereas removing non-evidence mass increases it by 0.0034 under an oracle counterfactual. In TKG, removing true-tail evidence mass reduces MRR by 0.1162, whereas removing non-evidence mass increases it by 0.2245 under an oracle counterfactual. In CTDG, removing counterpart-history mass reduces AUC by 0.0575, and removing non-evidence mass also reduces AUC, by 0.0203. Thus, surrounding temporal context is useful for CTDG edge prediction, although the larger dependency lies on the counterpart-evidence path. Permission accordingly acts as contribution scaling for the current query rather than binary evidence/non-evidence masking.

\paragraph{RAG case study.}
Figure~\ref{fig:rag_permission_case} presents a HotpotQA case study, showing raw attention, \method{} permission, and effective mass side by side on a passage--query grid. Raw attention \(\alpha_{ij}\) can assign mass to both supporting and distractor passages. Permission \(g_{ij}\) in the middle panel remains high at annotated supporting-passage cells and decreases at distractor cells attended through lexical or contextual relevance. The effective mass \(\alpha_{ij}g_{ij}\) in the right panel combines these signals: supporting cells remain in the prediction-facing support score, while distractor cells contribute less even with the same raw attention. This example illustrates that the RAG attention map is not itself an evidence map; permission reshapes the mass entering supporting-evidence ranking.

\begin{center}
\includegraphics[width=\textwidth]{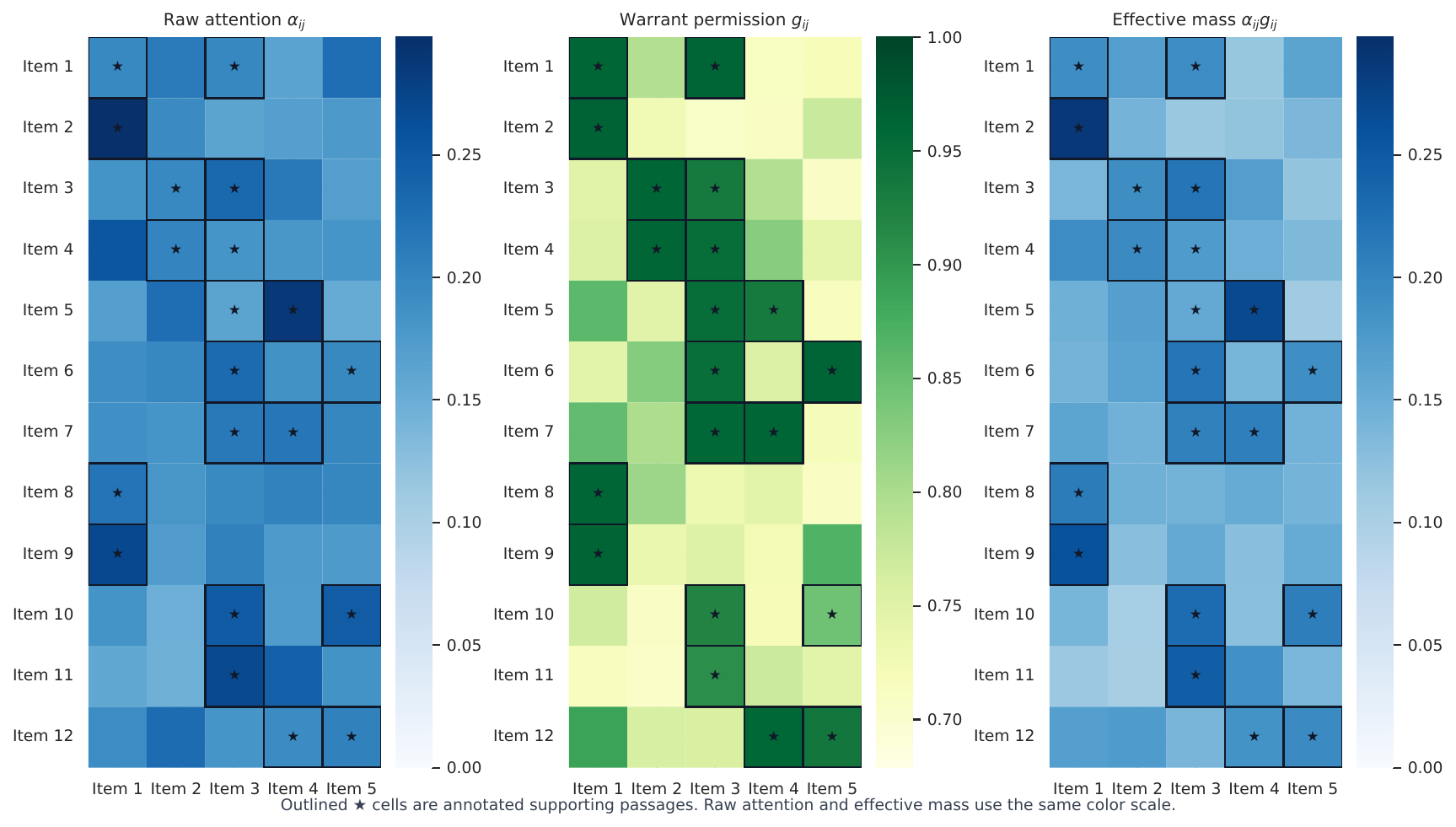}
\captionof{figure}{HotpotQA RAG case study. Left: raw attention \(\alpha_{ij}\). Middle: \method{} permission \(g_{ij}\). Right: effective mass \(\alpha_{ij}g_{ij}\). Outlines and stars mark annotated supporting-passage cells.}
\label{fig:rag_permission_case}
\end{center}

\subsection{Scope Conditions and Failure Regimes}
\label{subsec:failure_regimes}

The effect of Full Warrant varies across domains and backbones. Mean improvements are small for MTPP and RAG, and permission's marginal effect over OpenPath is negligible or negative in some CTDG and STPP rows. Permission is therefore evaluated through same-path controls and task-specific metrics.

On Gowalla/DeepSTPP, Full has higher Location RMSE than Base. On ICEWS18/CyGNet, Base obtains 0.1536 MRR, OpenPath 0.1438, and Full 0.1501. The ungated path is clearly harmful, and permission recovers 0.0063 MRR without fully surpassing Base. This is a partial-recovery condition: the additional improvement from learned permission does not fully restore the original backbone's performance.

\section{Discussion}

\label{sec:discussion}

\label{subsec:combined_statement}

\paragraph{Selective transmission rather than unrestricted access.}

Paths that read additional information and controls that use it beneficially serve different roles. In CyGNet, using the same connection without a gate reduced performance on all three datasets, while learning permission recovered all or part of the loss. In HotpotQA, learned gates suppressed distractors more strongly than gold support while improving ranking and unsupported attribution. These results show that reading query-relevant items does not require transmitting all their contributions to prediction at the same strength.

\paragraph{What the controls establish.}

OpenPath--Full comparisons measure the additional effect of learning permission on the same implemented path. They distinguish cases such as CTDG, where the overall Base--Full gain is large but the permission increment is small. Normalized, scalar, item-only, and attention-adapter controls examine relative allocation, total transmission mass, and conditioning. The most effective control varies by task. Full achieves higher means in the G1/G2 comparison, alongside seed variation and differences in implementation location. Thus, the results show task-dependent effects of control units and degrees of freedom rather than universal superiority of a particular gate form.

\paragraph{How the diagnostic evidence connects to learned control.}

Contribution-level diagnostics and learned-permission experiments answer different questions. Item-removal analysis shows that even highly attended contributions can reduce current prediction utility. Oracle mass interventions show that selectively removing some non-evidence contributions can improve the actual metric. These experiments diagnose opportunities to selectively reduce already read contributions rather than measure the performance of learned Warrant itself. In contrast, OpenPath--Full comparisons and the HotpotQA learned-gate analysis measure prediction changes when control is learned through the task objective. In particular, HotpotQA assigns lower permission to distractors than to gold support while improving both Support MRR and unsupported attribution. The experiments therefore provide complementary evidence at three levels: the mismatch between attention and prediction utility, opportunities for improvement through selective suppression, and the actual behavior of learned permission.

\paragraph{Scope of the evidence.}

Because OpenPath and Full are trained independently, joint changes in gate distributions and rankings do not isolate the causal effects of individual gate values. The G1/G2 comparison also includes differences in application layers and gate parameterization. The overall benchmark sign test assumes independent comparisons, whereas rows share datasets and backbones. RAG evaluation covers supporting evidence selection; answer generation and hallucination require separate evaluation.

\paragraph{Limitations.}

On ICEWS18/CyGNet, performance remains below Base after recovery, and in some settings Full also underperforms OpenPath. A scalar gate cannot separately control helpful and harmful components mixed within the same item, and small attention weights and sigmoid saturation can weaken learning signals. Jointly trained attention and gates also lack guaranteed semantic identifiability. Path construction requires access to the computation graph and prediction tensors, together with backbone-specific implementation. The local loss and SNR analyses apply under their stated assumptions and do not guarantee monotonic improvement during joint updates of shared parameters.

\section{Conclusion}
\label{sec:conclusion}

\method{} learns how strongly attention-derived item contributions should reach prediction under the current query. Permission is applied before aggregation and jointly controls relative allocation across items and total transmission mass. Same-path comparisons across five task families show effects that vary by task and backbone. In particular, permission recovers losses caused by ungated paths in CyGNet and improves evidence ranking and unsupported attribution in HotpotQA while suppressing distractors more strongly than gold support. These results demonstrate the utility and scope of learning contribution transmission strengths rather than equating attention relevance with prediction utility.

\printcredits
\section*{Declaration of competing interest}
The authors declare that they have no known competing financial interests or
personal relationships that could have appeared to influence the work reported
in this paper.

\section*{Acknowledgements}
The authors have no acknowledgements to declare.

\section*{Data availability}

All datasets used in this study are publicly available. The exact data sources are listed below.

\begin{center}
\small
\setlength{\tabcolsep}{4pt}
\renewcommand{\arraystretch}{1.15}
\begin{tabular}{p{0.12\textwidth}p{0.22\textwidth}p{0.58\textwidth}}
\toprule
Domain & Dataset & Source \\
\midrule
CTDG & Wikipedia & \url{https://snap.stanford.edu/jodie/wikipedia.csv} \\
CTDG & MOOC & \url{https://snap.stanford.edu/jodie/mooc.csv} \\
CTDG & LastFM & \url{https://snap.stanford.edu/jodie/lastfm.csv} \\
MTPP & StackOverflow & HuggingFace: \url{https://huggingface.co/datasets/easytpp/stackoverflow} \\
MTPP & Retweets & HuggingFace: \url{https://huggingface.co/datasets/easytpp/retweet} \\
STPP & Earthquake & GitHub: \url{https://github.com/ss15859/EarthquakeNPP} \\
STPP & Gowalla & \url{https://snap.stanford.edu/data/loc-gowalla_totalCheckins.txt.gz} \\
TKG & ICEWS14 & GitHub RE-Net data: \url{https://github.com/INK-USC/RE-Net} \\
TKG & ICEWS18 & GitHub RE-Net data: \url{https://github.com/INK-USC/RE-Net} \\
TKG & GDELT & GitHub RE-Net data: \url{https://github.com/INK-USC/RE-Net} \\
RAG & HotpotQA & HuggingFace: \url{https://huggingface.co/datasets/hotpotqa/hotpot_qa} \\
RAG & Natural Questions & GitHub: \url{https://github.com/google-research-datasets/natural-questions} \\
\bottomrule
\end{tabular}
\end{center}

\section*{Code availability}

The implementation, experiment configurations, aggregation scripts, and plotting scripts used in this study are available at \url{https://github.com/SnowyPainter/warrant-public}.

\bibliographystyle{cas-model2-names}
\bibliography{references}

\appendix
\section*{Appendix}

\counterwithin{figure}{section}
\counterwithin{table}{section}

\section{Theoretical Derivations}
\label{app:theoretical_derivations}

This appendix develops the mathematical structure of \method{} as a permission operator on metric-facing attention-weighted value terms. For a single query, we write
\[
\begin{aligned}
h &= \sum_j\alpha_jv_j, &
c_j &= \alpha_jv_j,\\
h^W &= \sum_j\alpha_jg_jv_j = \sum_jg_jc_j, &
g_j &= \lambda+(1-\lambda)\sigma(\psi_j).
\end{aligned}
\]
Here, \(\lambda>0\) is the leak factor. Thus, \method{} is not a hard mask: every valid item retains positive value scaling. This does not prevent gate gradients from diminishing through sigmoid saturation or small attention weights.

\subsection{Gate Gradient}
\label{app:gate_gradient}

Let the metric-facing loss be \(\mathcal{L}(h^W)\). Since \(\psi_j\) directly affects only \(g_j\),
\[
\begin{aligned}
\frac{\partial h^W}{\partial g_j} &= \alpha_jv_j = c_j,\\
\frac{\partial g_j}{\partial\psi_j} &= (1-\lambda)\sigma'(\psi_j),\\
\frac{\partial\mathcal{L}}{\partial \psi_j}
&= (1-\lambda)\sigma'(\psi_j)
\left\langle \nabla_{h^W}\mathcal{L},c_j\right\rangle .
\end{aligned}
\]
By the chain rule, the gate receives an item-wise, loss-aligned signal from the weighted-value term itself. A positive inner product indicates that the contribution points toward increasing loss, so gradient descent lowers its permission logit. A negative inner product indicates a loss-reducing direction, moving the gate toward preservation.

\subsection{Local Loss Bound}
\label{app:local_loss_bound}

Suppose \(\mathcal{L}\) is \(\beta\)-smooth with respect to the prediction representation. With the ungated reference state \(h=\sum_jc_j\), scaling only item \(j\) gives
\[
h(g_j)=h+(g_j-1)c_j.
\]
Smoothness yields the following bound.
\[
\begin{aligned}
\mathcal{L}(h(g_j))
&\le \mathcal{L}(h)
+(g_j-1)\left\langle\nabla\mathcal{L}(h),c_j\right\rangle
+\frac{\beta}{2}(g_j-1)^2\|c_j\|^2,\\
g_j^\star
&= \Pi_{[\lambda,1]}
\left(
1-
\frac{\left\langle\nabla\mathcal{L}(h),c_j\right\rangle}{\beta\|c_j\|^2}
\right).
\end{aligned}
\]
For $c_j\ne0$ and $\beta>0$, this expression is the one-dimensional minimizer of the upper bound. It selects attenuation for positive gradient alignment and $g_j=1$ for negative alignment. If $c_j=0$, the gate does not affect the output. A gate trained through the task loss need not attain this bound-minimizing value.

\subsection{Manifold-Local Permission Rule}
\label{app:manifold_local_permission}

The Euclidean local rule in Appendix~\ref{app:local_loss_bound} is formulated in the ambient representation space. Suppose the prediction state lies near a local \(C^2\) representation manifold \(\mathcal{M}\subset\mathbb{R}^d\), with tangent space \(T_h\mathcal{M}\) and tangent projection \(P_h\). Writing \(g_j=1-\delta_j\), the \(\method{}\) perturbation is
\[
\Delta(g)=\sum_j(g_j-1)c_j=-\sum_j\delta_jc_j
\]
The component that directly enters the first-order loss change on the manifold is
\[
\xi(g)=P_h\Delta(g)=-\sum_j\delta_jP_hc_j
\]
Let \(\tilde c_j=P_hc_j\). Riemannian smoothness for a local retraction \(R_h\) gives
\[
\mathcal{L}(R_h(\xi))
\le
\mathcal{L}(h)
+
\langle \operatorname{grad}_{\mathcal{M}}\mathcal{L}(h),\xi\rangle
+
\frac{\beta_{\mathcal{M}}}{2}\|\xi\|^2
\]
When only one contribution is controlled, \(\xi=-\delta_j\tilde c_j\), and the upper bound excluding its constant term is
\[
-\delta_j
\langle \operatorname{grad}_{\mathcal{M}}\mathcal{L}(h),\tilde c_j\rangle
+
\frac{\beta_{\mathcal{M}}}{2}\delta_j^2\|\tilde c_j\|^2
\]
For $\|\tilde c_j\|>0$, the unconstrained minimizer is
\[
\delta_j=
\frac{
\langle \operatorname{grad}_{\mathcal{M}}\mathcal{L}(h),\tilde c_j\rangle
}{
\beta_{\mathcal{M}}\|\tilde c_j\|^2
}
\]
Projection onto the feasible interval \([0,1-\lambda]\) yields the local projected rule. An optional \(\epsilon\) in numerical computation stabilizes the denominator when \(\|\tilde c_j\|\) is near zero; it is not part of the exact minimizer. This result describes local tangent-space descent rather than global optimality.

\subsection{Tangent Gram Matrix and Entangled Paths}
\label{app:tangent_gram_entanglement}

When multiple contributions are controlled jointly, \(\xi(g)=-\sum_j\delta_j\tilde c_j\). Define
\[
 a_j=\langle \operatorname{grad}_{\mathcal{M}}\mathcal{L}(h),\tilde c_j\rangle,
 \qquad
 G_{jk}=\langle\tilde c_j,\tilde c_k\rangle .
\]
Substituting into the Riemannian smoothness bound gives
\[
\begin{aligned}
\mathcal{L}(R_h(\xi(g)))
&\le
\mathcal{L}(h)
-
\sum_j\delta_ja_j
+
\frac{\beta_{\mathcal{M}}}{2}
\left\|\sum_j\delta_j\tilde c_j\right\|^2 \\
&=
\mathcal{L}(h)
-
\delta^\top a
+
\frac{\beta_{\mathcal{M}}}{2}\delta^\top G\delta .
\end{aligned}
\]
Here, \(G\) is the Gram matrix of tangent contributions. Off-diagonal entries \(G_{jk}\) represent coupling between contributions and can be positive or negative. Positive coupling increases the upper bound, whereas negative coupling decreases it. Rules that control each item independently therefore do not generally capture interactions among contributions. When helpful and harmful terms are directionally entangled, the effect of scalar permission also depends on the transmission strengths of other items.

\subsection{Smooth Bounded Parameterization and Stationary Scope}
\label{app:bounded_optimization_scope}

The leak-sigmoid gate is defined as
\[
 g_j=\lambda+(1-\lambda)\sigma(\psi_j)
\]
It therefore satisfies
\[
 g_j\in[\lambda,1],
 \qquad
 \left|\frac{\partial g_j}{\partial\psi_j}\right|
 =(1-\lambda)\sigma(\psi_j)(1-\sigma(\psi_j))
 \le \frac{1-\lambda}{4}.
\]
Furthermore, since \(\delta_j=1-g_j\in[0,1-\lambda]\),
\[
\|\Delta(g)\|
=
\left\|\sum_j(g_j-1)c_j\right\|
\le
\sum_j\delta_j\|c_j\|
\le
(1-\lambda)\sum_j\|c_j\| .
\]
Thus, \method{} is a smooth, bounded perturbation of the base value path rather than a hard discontinuous mask. Assume that the augmented objective $F(\theta)$ is $L_F$-smooth and bounded below over the region containing the iterates, and that stochastic gradients are unbiased with bounded variance. Bounded individual gates alone do not ensure these assumptions for the full objective. For a step size \(\eta\le 1/L_F\), the standard descent lemma yields
\[
\frac{1}{T}\sum_{t=0}^{T-1}
\mathbb{E}\|\nabla F(\theta_t)\|^2
\le
\frac{2(F(\theta_0)-F_\star)}{\eta T}
+
L_F\eta\sigma_g^2
\]
This is a convergence-to-stationarity bound. Its scope is that bounded, smooth parameterization preserves ordinary smooth trainability under the stated assumptions, rather than a \method{}-specific guarantee of global convergence or improvement over the base model.

\subsection{SNR Condition}
\label{app:snr_condition}

We analyze one scalar prediction direction with fixed attention and gates. Assume a positive sum of support signals, mutually uncorrelated noise terms, and a positive sum of noise variances. Decompose contributions into support signals and zero-mean noise:
\[
\begin{aligned}
h_B=
\sum_{j\in S}\alpha_j\mu_j+
\sum_{j\in N}\alpha_j\xi_j,\qquad
\mathbb{E}[\xi_j] &= 0,
\quad
\mathrm{Var}(\xi_j)=\sigma_j^2,\\
\mathrm{SNR}_B=
\frac{\sum_{j\in S}\alpha_j\mu_j}{\sqrt{\sum_{j\in N}\alpha_j^2\sigma_j^2}},\qquad
\mathrm{SNR}_W=
\frac{\sum_{j\in S}\alpha_jg_j\mu_j}{\sqrt{\sum_{j\in N}\alpha_j^2g_j^2\sigma_j^2}}.
\end{aligned}
\]
The SNRs of Base and \method{} are \(\mathrm{SNR}_B\) and \(\mathrm{SNR}_W\), respectively. Support-signal retention and noise-standard-deviation retention are
\[
\begin{aligned}
R_S=
\frac{\sum_{j\in S}\alpha_jg_j\mu_j}{\sum_{j\in S}\alpha_j\mu_j},\qquad
R_N=
\sqrt{
\frac{\sum_{j\in N}\alpha_j^2g_j^2\sigma_j^2}{\sum_{j\in N}\alpha_j^2\sigma_j^2}
},\\
\frac{\mathrm{SNR}_W}{\mathrm{SNR}_B}
&=\frac{R_S}{R_N},\qquad
\mathrm{SNR}_W>\mathrm{SNR}_B
\Longleftrightarrow R_S>R_N.
\end{aligned}
\]
This condition does not require perfect preservation of support signals. SNR increases when support signals are retained more strongly than the noise standard deviation.

\subsection{High-Probability Interpretation}
\label{app:high_probability_snr}

The quantities in the SNR condition are weighted averages. Define the weights
\[
\begin{aligned}
w_j^S=\frac{\alpha_j\mu_j}{\sum_{l\in S}\alpha_l\mu_l},
w_j^N=\frac{\alpha_j^2\sigma_j^2}{\sum_{l\in N}\alpha_l^2\sigma_l^2},\\
R_S&=\sum_{j\in S}w_j^Sg_j,
&
R_N^2&=\sum_{j\in N}w_j^Ng_j^2.
\end{aligned}
\]
Since \(g_j\in[\lambda,1]\), both quantities are bounded weighted averages. Under fixed nonnegative weights, independent bounded gate sampling, and a positive expectation margin between the weighted support mean and the noisy-gate RMS, weighted concentration makes the false-suppression event \(R_S\le R_N\) less likely as the effective support/noise item counts increase. This interpretation depends on the sampling assumptions and does not directly extend to strongly dependent gates learned from a shared query.

\subsection{Gradient Noise and Effective Curvature}
\label{app:gradient_noise_curvature}

In a conditional, path-local analysis with fixed attention and gates, let \(\xi_j\) denote a noisy value-gradient component. The Base path transmits \(\alpha_j\xi_j\), while the \method{} path transmits \(\alpha_jg_j\xi_j\). Hence,
\[
\begin{aligned}
\mathrm{Var}[\alpha_jg_j\xi_j]
&=\alpha_j^2g_j^2\mathrm{Var}[\xi_j]
\le \alpha_j^2\mathrm{Var}[\xi_j],\\
\|\nabla_{v_j}^2\mathcal{L}\|_{\mathrm{base}}
&\le \alpha_j^2\beta,
&
\|\nabla_{v_j}^2\mathcal{L}\|_{\method{}}
&\le \alpha_j^2g_j^2\beta
\le \alpha_j^2\beta.
\end{aligned}
\]
Similarly, if \(\mathcal{L}\) is \(\beta\)-smooth in \(h\), the second line holds along item value \(v_j\). These are path-local bounds: down-weighting the value path lowers the noise and curvature bounds under these conditions. They do not bound the full model Hessian or optimizer variance when derivatives of the gates and attention are included.

\subsection{Diagonal Permission vs. Softmax Reweighting}
\label{app:diagonal_permission}

Let \(m_j^W=\alpha_jg_j\). Since \method{} does not recompute the attention softmax,
\[
\begin{aligned}
\frac{\partial m_j^W}{\partial\psi_l}
&=\alpha_j(1-\lambda)\sigma'(\psi_j)\mathbf{1}[j=l],\\
\tilde{\alpha}_j
&=\mathrm{softmax}_j(s_j+\log g_j),\\
\frac{\partial\tilde{\alpha}_j}{\partial\log g_l}
&=\tilde{\alpha}_j(\mathbf{1}[j=l]-\tilde{\alpha}_l).
\end{aligned}
\]
The first derivative has an item-wise diagonal structure. In contrast, attention-logit gating couples all items through softmax normalization. Thus, \method{} scales contributions while preserving relevance \(\alpha_j\), whereas logit gating changes the relevance distribution itself.

\subsection{Item-Wise and Post-Aggregation Control}
\label{app:post_attention_path}

Consider a controller \(F(h)\) that receives only the aggregate \(h=\sum_j\alpha_jv_j\). For the same query, item configurations \((u,-u)\) and \((2u,-2u)\) with weights \((.5,.5)\) both yield \(h=0\). Applying the same gates \((a,b)\), with \(a\ne b\), produces item-wise outputs \(.5(a-b)u\) and \((a-b)u\), respectively. Since \(F(0)\) returns the same value for both configurations, it cannot represent both outputs simultaneously. This distinguishes a function observing only the aggregate from control that retains item-level information.

Attention-logit gates can retain item identity but redistribute attention mass through softmax normalization: lowering one item automatically increases the mass of others. \method{} uses \(m_j^W=\alpha_jg_j\) without enforcing \(\sum_jm_j^W=1\). This operation attenuates weighted-value contributions rather than redistributing relevance.

\begin{center}
\begin{minipage}{\textwidth}
\centering
\captionof{table}{Same-aggregate constructive check. This toy experiment isolates function-class differences arising from the gate's control target.}
\label{tab:same_aggregate_constructive_check}
\scriptsize
\setlength{\tabcolsep}{4pt}
\renewcommand{\arraystretch}{1.12}
\resizebox{\textwidth}{!}{
\begin{tabular}{lllll}
\toprule
Variant & Item identity available & Attention renormalized & Expected accuracy & Interpretation \\
\midrule
\texttt{post\_attention\_gate} & no & no & 0.5 & aggregate gate sees the same zero vector for both labels \\
\texttt{attention\_logit\_gate} & yes & yes & 1.0 & separates items by redefining attention mass \\
\texttt{warrant\_value\_term\_gate} & yes & no & 1.0 & keeps \(\alpha\) fixed and changes item-wise weighted value contribution \\
\bottomrule
\end{tabular}
}
\end{minipage}
\end{center}

In contrast, \method{} is applied before item identity is lost.
\[
\begin{aligned}
h^W &= \sum_j\alpha_jg(q,k_j)v_j,\\
\frac{\partial\mathcal{L}}{\partial\psi_j}
&= (1-\lambda)\sigma'(\psi_j)
\left\langle
J_p^\top\nabla_z\mathcal{L},
\alpha_jv_j
\right\rangle,
&
J_p &= \frac{\partial z}{\partial h_p}.
\end{aligned}
\]
Here, the gated path is \(h_p=\sum_j\alpha_jg_jv_j\), and the metric object is \(z=f(h_p)\). The gate's learning signal is proportional to the path-to-metric Jacobian. Path sensitivity to prediction therefore affects the magnitude of the permission learning signal. Placing \method{} on a path weakly coupled to the metric weakens that signal, whereas correct-path placement exposes weighted-value terms that actually change the reported metric.

\section{Monte Carlo Details}
\label{app:monte_carlo_details}

Figure~\ref{fig:snr_improvement_region} visualizes the SNR condition derived in Appendix~\ref{app:snr_condition}. The left panel shows the deterministic boundary \(R_S=R_N\), where \(R_S\) and \(R_N\) are the support-signal and noise-standard-deviation retention factors defined in Appendix~\ref{app:snr_condition}. The right panel samples the same quantities under evidence-aligned gates. The weighted-average interpretation follows Appendix~\ref{app:high_probability_snr}.

\begin{center}
\includegraphics[width=0.86\textwidth]{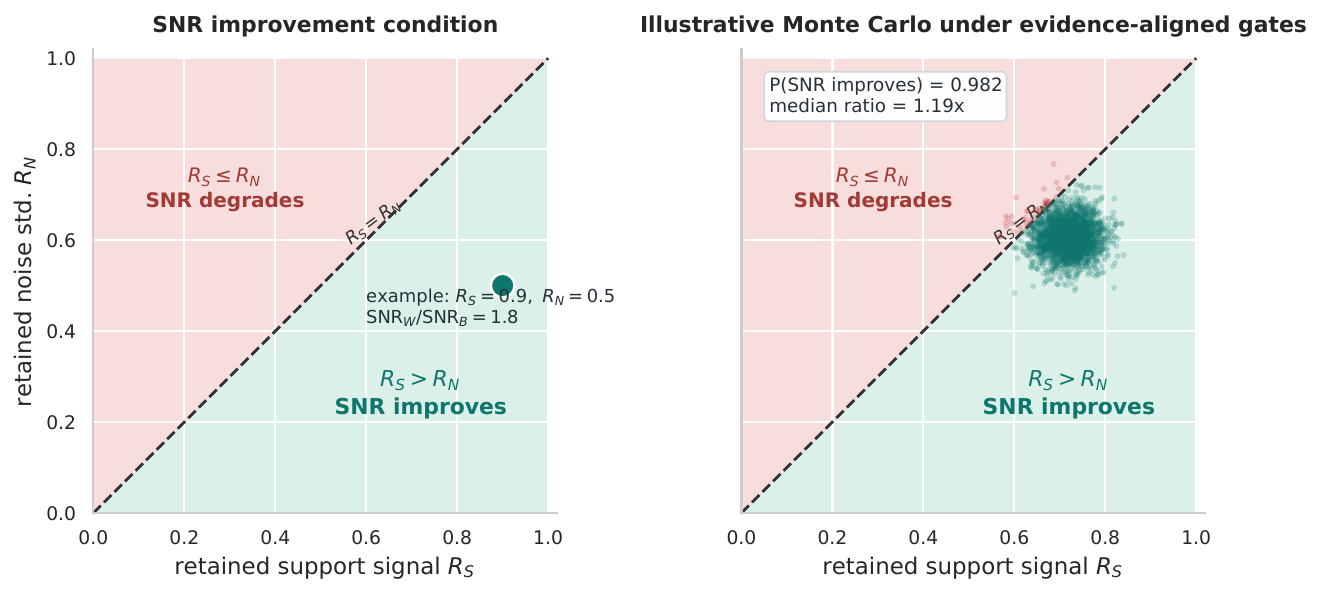}
\captionof{figure}{SNR improvement condition under evidence-aligned permission. The diagonal boundary is \(R_S=R_N\); sampled operating points use the Monte Carlo setting described below.}
\label{fig:snr_improvement_region}
\end{center}

Each trial samples a support set \(S\) and a noisy-item set \(N\), with \(|S|=8\) support items and \(|N|=32\) noisy items. Attention weights for support and noisy items are independently sampled from normalized Gamma variables to form random simplex weights. Support magnitudes \(\mu_j\) and noise standard deviations \(\sigma_j\) are sampled from log-normal distributions to avoid a degenerate equal-weight setting.
\[
\begin{aligned}
\mu_j &\sim \mathrm{LogNormal}(0,0.25^2), &
\sigma_j &\sim \mathrm{LogNormal}(0,0.25^2).
\end{aligned}
\]
Gates are sampled from Beta distributions. In the evidence-aligned setting of Figure~\ref{fig:snr_improvement_region}, the support gate mean is \(0.72\), the noisy gate mean is \(0.60\), and both distributions have concentration parameter \(30\).
\[
\begin{aligned}
g_j^S &\sim \mathrm{Beta}(0.72\cdot 30,(1-0.72)\cdot 30),\\
g_j^N &\sim \mathrm{Beta}(0.60\cdot 30,(1-0.60)\cdot 30).
\end{aligned}
\]
This is a weak-alignment setting rather than a near-perfect separator. Support gates are only moderately larger than noisy gates, and some sampled points remain in the degradation region.

For each trial, we compute \(R_S\), \(R_N\), and \(\mathrm{SNR}_W/\mathrm{SNR}_B=R_S/R_N\) as defined in Appendix~\ref{app:snr_condition}. A trial improves when \(R_S>R_N\). We run 2,500 trials with a fixed random seed. Under this weak evidence alignment, \(98.2\%\) of sampled operating points satisfy \(R_S>R_N\), and the median SNR ratio is \(1.19\times\). Lowering support gates below noisy gates shifts sampled points toward the degradation region in Appendix~\ref{app:failure_regimes}.

\section{Manifold-Local Simulation}
\label{app:manifold_local_simulation}

This appendix presents a synthetic decomposition of the tangent Gram bound in Appendix~\ref{app:tangent_gram_entanglement}. The illustrative construction shows how the cross term in \(\delta^\top G\delta\) can reduce the benefit of diagonal shrinkage. For two tangent contributions, set \(a=(1,1)\), \(\delta=(0.5,0.5)\), and
\[
G=\begin{bmatrix}1&\rho\\\rho&1\end{bmatrix}
\]
The descent term is then \(\delta^\top a=1.0\), and the bound change is
\[
\Delta_{\mathrm{bound}}
=
-\delta^\top a
+
\frac{\beta_{\mathcal{M}}}{2}\delta^\top G\delta
\]
Table~\ref{tab:manifold_local_simulation} shows that the cross term increases with \(\rho\) and curvature \(\beta_{\mathcal{M}}\); the bound can become positive in the high-curvature, entangled case.

\begin{center}
\begin{minipage}{\textwidth}
\centering
\captionof{table}{Synthetic manifold-local bound decomposition. The table is an illustrative instantiation of Appendix~\ref{app:tangent_gram_entanglement}, not a benchmark result. Larger off-diagonal tangent coupling increases the cross term and can overturn the diagonal shrinkage benefit under high curvature.}
\label{tab:manifold_local_simulation}
\scriptsize
\setlength{\tabcolsep}{4pt}
\renewcommand{\arraystretch}{1.12}
\begin{tabular}{lrrrrrr}
\toprule
Regime
& $\beta_{\mathcal M}$
& $\rho$
& $-\delta^\top a$
& diagonal term
& cross term
& $\Delta_{\mathrm{bound}}$ \\
\midrule
Low entanglement
& 1.0 & 0.05 & $-1.000$ & 0.250 & 0.013 & $-0.738$ \\
Moderate entanglement
& 1.0 & 0.40 & $-1.000$ & 0.250 & 0.100 & $-0.650$ \\
High entanglement
& 1.0 & 0.80 & $-1.000$ & 0.250 & 0.200 & $-0.550$ \\
High-curvature entangled
& 3.0 & 0.80 & $-1.000$ & 0.750 & 0.600 & $+0.350$ \\
\bottomrule
\end{tabular}
\end{minipage}
\end{center}

This simulation supports the failure-condition interpretation in the main text. Even with high path sensitivity, scalar item-wise permission may fail to improve prediction safely when useful and shortcut/noisy contributions are strongly entangled in the tangent space.

\section{Failure Regimes}
\label{app:failure_regimes}

The analysis identifies three failure conditions. First, \emph{false suppression} occurs when support items receive low gates such that \(R_S\le R_N\). Second, \emph{unaligned permission} occurs when noisy or shortcut items receive gates as high as support items, leaving noise retention unreduced. Third, \emph{weak prediction coupling} occurs when \method{} is placed on a path with a small path-to-metric Jacobian, providing only a weak metric-facing learning signal. We therefore evaluate primary metrics alongside support/distractor mass, gate ratios, and correct-path versus generic-placement diagnostics.

\section{Additional HotpotQA/RoBERTa Details}
\label{app:hotpotqa_roberta}

This appendix provides detailed settings, complete per-variant results, and permission diagnostics for the pretrained RoBERTa supporting sentence selection controls in Section~\ref{subsec:pretrained_transformer_permission}. It supplements the input construction, training settings, readout variants, and seed-wise means and standard deviations.

\begin{center}
\begin{minipage}{\textwidth}
\centering
\captionof{table}{HotpotQA/RoBERTa candidate-marker control settings. This is an encoder-side evidence-attribution task ranking candidate sentences as annotated supporting evidence, rather than evaluating answer generation.}
\label{tab:hotpotqa_roberta_setup}
\scriptsize
\setlength{\tabcolsep}{4pt}
\renewcommand{\arraystretch}{1.12}
\resizebox{\textwidth}{!}{
\begin{tabular}{lll}
\toprule
Group & Component & Setting \\
\midrule
Encoder & base model & \texttt{roberta-base} \\
Data & dataset & HotpotQA processed supporting sentence candidates \\
Input & format & question + instruction + \(\langle C0\rangle,\ldots,\langle C7\rangle\) candidate evidence sentences \\
Input & candidate count & maximum 8 candidates per question \\
Input & candidate composition & supporting candidates + hard distractors + random distractors \\
Training & questions / seeds / epochs & maximum 2,000 questions; seeds \(7,17,37\); 2 epochs \\
Training & main loss & candidate-wise binary cross entropy \\
Training & permission auxiliary & weak pairwise gate-alignment auxiliary loss \\
Evaluation & primary metric & Support MRR over candidate ranking \\
Evaluation & unsupported attribution & unsupported candidate selected within gold-count budget \\
\midrule
Control & \texttt{base} & final candidate marker readout baseline \\
Control & \texttt{param\_mlp} & parameter-count control without value-term permission \\
Control & \texttt{post\_attention\_glu} & aggregate/readout gate after item identity is mixed \\
Control & \texttt{attention\_readout} & ungated candidate-marker value-path exposure \\
Control & \texttt{query\_only\_gate} & generic permission without item key conditioning \\
Control & \texttt{shuffled\_warrant} & query-item pairing is broken \\
Control & \texttt{openpath\_nogate} & same metric-facing path with \(g=1\) \\
Control & \texttt{full\_warrant} & \(\alpha_{ij}v_j\mapsto\alpha_{ij}g_{ij}v_j\) on the candidate-marker value path \\
\bottomrule
\end{tabular}
}
\end{minipage}
\end{center}

\begin{center}
\begin{minipage}{\textwidth}
\centering
\captionof{table}{Complete HotpotQA/RoBERTa supporting sentence selection controls. Values are means and standard deviations over three seeds. Bold indicates the best value for each metric.}
\label{tab:hotpotqa_roberta_results}
\scriptsize
\setlength{\tabcolsep}{3.0pt}
\renewcommand{\arraystretch}{1.12}
\resizebox{\textwidth}{!}{
\begin{tabular}{lrrrrr}
\toprule
Variant & MRR \(\uparrow\) & R@1 \(\uparrow\) & Evidence F1 \(\uparrow\) & AUPRC \(\uparrow\) & Unsupported@GoldCount \(\downarrow\) \\
\midrule
\texttt{base} & \(0.9111 \pm 0.0045\) & \(0.8371 \pm 0.0073\) & \(0.7751 \pm 0.0049\) & \(0.8110 \pm 0.0065\) & \(0.2249 \pm 0.0049\) \\
\texttt{param\_mlp} & \(0.9036 \pm 0.0057\) & \(0.8245 \pm 0.0106\) & \(0.7688 \pm 0.0033\) & \(0.8103 \pm 0.0082\) & \(0.2312 \pm 0.0033\) \\
\texttt{post\_attention\_glu} & \(0.9051 \pm 0.0039\) & \(0.8295 \pm 0.0098\) & \(0.7629 \pm 0.0084\) & \(0.8034 \pm 0.0116\) & \(0.2371 \pm 0.0084\) \\
\texttt{attention\_readout} & \(0.9125 \pm 0.0042\) & \(0.8405 \pm 0.0072\) & \(0.7730 \pm 0.0076\) & \(\mathbf{0.8129} \pm 0.0040\) & \(0.2270 \pm 0.0076\) \\
\texttt{query\_only\_gate} & \(0.9078 \pm 0.0021\) & \(0.8312 \pm 0.0048\) & \(0.7726 \pm 0.0057\) & \(0.8104 \pm 0.0053\) & \(0.2274 \pm 0.0057\) \\
\texttt{shuffled\_warrant} & \(0.9046 \pm 0.0102\) & \(0.8270 \pm 0.0166\) & \(0.7671 \pm 0.0160\) & \(0.7979 \pm 0.0086\) & \(0.2329 \pm 0.0160\) \\
\texttt{openpath\_nogate} & \(0.9129 \pm 0.0092\) & \(0.8405 \pm 0.0165\) & \(0.7755 \pm 0.0110\) & \(0.8102 \pm 0.0127\) & \(0.2245 \pm 0.0110\) \\
\texttt{full\_warrant} & \(\mathbf{0.9134} \pm 0.0010\) & \(\mathbf{0.8414} \pm 0.0052\) & \(\mathbf{0.7772} \pm 0.0058\) & \(0.8042 \pm 0.0027\) & \(\mathbf{0.2228} \pm 0.0058\) \\
\bottomrule
\end{tabular}
}
\end{minipage}
\end{center}

\begin{center}
\begin{minipage}{\textwidth}
\centering
\captionof{table}{HotpotQA/RoBERTa permission diagnostics. Gold/Random ratio denotes the average mass ratio between annotated supporting candidates and random distractors.}
\label{tab:hotpotqa_roberta_diagnostics}
\scriptsize
\setlength{\tabcolsep}{4pt}
\renewcommand{\arraystretch}{1.12}
\resizebox{\textwidth}{!}{
\begin{tabular}{lrrrrr}
\toprule
Variant & Gate mean & Gold gate & Random gate & Gold/Random \(\alpha\) & Gold/Random \(\alpha g\) \\
\midrule
\texttt{attention\_readout} & \(0.3671\) & \(0.8688\) & \(0.1745\) & \(0.3811\) & \(2.2536\) \\
\texttt{query\_only\_gate} & \(0.5416\) & \(0.8610\) & \(0.4161\) & \(0.9987\) & \(\mathbf{5.0091}\) \\
\texttt{shuffled\_warrant} & \(0.9994\) & \(0.9994\) & \(0.9994\) & \(0.9013\) & \(0.9015\) \\
\texttt{openpath\_nogate} & \(1.0000\) & \(1.0000\) & \(1.0000\) & \(0.6265\) & \(0.6265\) \\
\texttt{full\_warrant} & \(0.5319\) & \(0.8647\) & \(0.4049\) & \(0.8179\) & \(3.8522\) \\
\bottomrule
\end{tabular}
}
\end{minipage}
\end{center}

\section{Prediction-Path Implementation}
\label{subsec:path_localization}

Permission is applied at the interface through which item-level contributions reach prediction. We identify the prediction tensor and upstream item axis in the backbone's computation graph and construct the required connection. Algorithm~\ref{alg:metric_path_tracing} provides a tracing protocol for checking this implementation. Backbone-specific connections are determined from the prediction tensor corresponding to the reported metric.

\begin{center}
\refstepcounter{algorithm}\label{alg:metric_path_tracing}
\fbox{%
\begin{minipage}{0.94\textwidth}
\small
\textbf{Algorithm~\thealgorithm. Prediction-path construction for \method{}}\\[2pt]
\textbf{Input:} base model \(F\), evaluation batch \(\mathcal{B}\), reported primary metric \(m\), candidate attention modules \(\mathcal{A}\), candidate prediction objects \(\mathcal{Z}\).\\
\textbf{Output:} metric-facing weighted value path \(p^\star\).
\begin{enumerate}
  \item Identify the prediction object \(z\in\mathcal{Z}\) from which \(m\) is computed, e.g., edge score, candidate mark logit, support logit, next-location state, tail score, or copy mass.
  \item Instrument candidate attention modules. For each module, expose attention weights \(\alpha_{ij}\), values \(v_j\), and weighted value terms \(c_{ij}=\alpha_{ij}v_j\).
  \item Enumerate candidate item-wise paths \(p\). Keep only paths where \(c_{ij}^{(p)}\) has an item axis \(j\) and lies upstream of \(z\) in the computation graph.
  \item Estimate path reach by Jacobian or intervention:
  \[
  \operatorname{Reach}(p)=
  \mathbb{E}_{\mathcal{B}}
  \frac{\|J_{z\leftarrow c^{(p)}}c^{(p)}\|}{\|z\|+\epsilon},
  \qquad
  \operatorname{Reach}_{0}(p)=
  \mathbb{E}_{\mathcal{B}}d\bigl(z,z[c^{(p)}\leftarrow0]\bigr).
  \]
  Discard paths whose perturbation changes hidden states but not the metric object \(z\).
  \item Select the last item-wise bottleneck before \(z\) among paths with non-negligible reach. Prefer the path closest to \(z\) that still preserves \(c_{ij}=\alpha_{ij}v_j\).
  \item Construct \method{} on \(p^\star\): \(\alpha_{ij}v_j\mapsto\alpha_{ij}g(q_i,k_j)v_j\).
  \item Validate the placement with Base vs. PathExposureOnly, Correct-path vs. Generic q-k \method{}, and Correct-path vs. Shuffled-pairing \method{} diagnostics.
\end{enumerate}
\end{minipage}}
\end{center}

Table~\ref{tab:tensor_path_localization} lists the actual tensor connections for each task. The protocol checks path reachability and item-wise controllability.

\begin{center}
\begin{minipage}{\textwidth}
\centering
\captionof{table}{Backbone-specific tensor connections for permission. Algorithm~\ref{alg:metric_path_tracing} checks reachability to the prediction object and preservation of the item axis.}
\label{tab:tensor_path_localization}
\scriptsize
\resizebox{\textwidth}{!}{
\begin{tabular}{lllll}
\toprule
Domain & Gated tensor & Query interface & Insertion point & Gradient path to metric \\
\midrule
CTDG & source/destination history term & current edge query & edge-score history aggregation & AUC edge score \\
MTPP & past event value term & candidate mark query & candidate mark residual logit & Mark MRR candidate score \\
RAG & retrieved passage value term & question-passage query & support ranking logit & Support MRR passage score \\
STPP & spatio-temporal history term & next-event dynamics query & next-location state update & Location RMSE dynamics state \\
TKG & historical fact tail term & \((h,r,t)\) query & tail score / copy path & Tail MRR ranking score \\
\bottomrule
\end{tabular}
}
\end{minipage}
\end{center}

Figure~\ref{fig:main_message} illustrates the implementation relationship between path construction and permission. OpenPath and Full share the same connection and differ in whether permission is learned. GenericGate controls for application placement.

\begin{center}
\resizebox{\textwidth}{!}{%
\definecolor{attnBlue}{HTML}{174EA6}
\definecolor{softBlue}{HTML}{EEF5FF}
\definecolor{warrantGreen}{HTML}{177A2E}
\definecolor{softGreen}{HTML}{EFF8EF}
\definecolor{warnOrange}{HTML}{D9471E}
\definecolor{softOrange}{HTML}{FFF4EC}
\definecolor{grayLine}{HTML}{6B6B6B}
\definecolor{softGray}{HTML}{F5F5F5}
\definecolor{darkText}{HTML}{1F1F1F}

\tikzset{
  >={Latex[length=2.5mm,width=2mm]},
  panelA/.style={
    draw=attnBlue,
    fill=softBlue,
    rounded corners=5pt,
    line width=0.9pt
  },
  panelB/.style={
    draw=warrantGreen,
    fill=softGreen,
    rounded corners=5pt,
    line width=0.9pt
  },
  panelC/.style={
    draw=grayLine,
    fill=softGray,
    rounded corners=5pt,
    line width=0.9pt
  },
  panelTitle/.style={
    font=\bfseries\normalsize,
    anchor=north west,
    inner sep=6pt
  },
  itemBox/.style={
    draw=attnBlue,
    fill=white,
    rounded corners=3pt,
    line width=0.7pt,
    minimum width=4.0cm,
    minimum height=1.15cm
  },
  queryBox/.style={
    draw=attnBlue,
    fill=white,
    rounded corners=4pt,
    line width=0.8pt,
    minimum width=1.45cm,
    minimum height=1.55cm,
    align=center,
    font=\small
  },
  formulaBoxBlue/.style={
    draw=attnBlue,
    fill=white,
    rounded corners=4pt,
    line width=0.8pt,
    minimum width=3.9cm,
    minimum height=1.0cm,
    align=center
  },
  warrantBox/.style={
    draw=warrantGreen,
    fill=white,
    rounded corners=4pt,
    line width=0.8pt,
    minimum width=2.15cm,
    minimum height=1.65cm,
    align=center,
    text width=1.9cm,
    inner sep=4pt,
    font=\small
  },
  bigFormulaGreen/.style={
    draw=warrantGreen,
    fill=white,
    rounded corners=4pt,
    line width=0.8pt,
    minimum width=8.2cm,
    minimum height=1.0cm,
    align=center
  },
  badge/.style={
    draw=warrantGreen,
    fill=white,
    rounded corners=4pt,
    line width=0.8pt,
    minimum height=0.72cm,
    align=center,
    font=\small,
    inner sep=3pt
  },
  grayCompBox/.style={
    draw=grayLine,
    fill=white,
    rounded corners=4pt,
    line width=0.75pt,
    minimum width=5.4cm,
    minimum height=1.18cm,
    align=center,
    text width=5.0cm,
    inner sep=4pt
  },
  taskBox/.style={
    draw=warrantGreen,
    fill=white,
    rounded corners=4pt,
    line width=0.8pt,
    minimum width=4.70cm,
    minimum height=1.28cm,
    align=center,
    text width=4.3cm,
    inner sep=4pt,
    font=\small
  },
  attnArrow/.style={
    ->,
    draw=attnBlue,
    line width=1.05pt
  },
  blackArrow/.style={
    ->,
    draw=darkText,
    line width=1.0pt
  },
  greenArrow/.style={
    ->,
    draw=warrantGreen,
    line width=1.0pt
  },
  dashedGreen/.style={
    draw=warrantGreen,
    dashed,
    line width=0.8pt
  },
  dashedGrayArrow/.style={
    ->,
    draw=grayLine,
    dashed,
    line width=0.9pt
  }
}

\begin{tikzpicture}[x=1cm,y=1cm,font=\sffamily]


\draw[panelA] (0, 0) rectangle (7.45, -10.25);
\draw[panelB] (7.65, 0) rectangle (20.25, -10.25);
\draw[panelC] (20.45, 0) rectangle (27.05, -10.25);
\draw[panelB] (0, -10.60) rectangle (27.05, -16.35);

\node[panelTitle, text=attnBlue] at (0, 0) {\textbf{a} \  Attention Access};

\node[queryBox] (query) at (1.05, -4.30) {query\\[-1pt]$q_i$};

\node[itemBox] (item1) at (4.90, -1.30) {};
\node[itemBox] (item2) at (4.90, -2.95) {};
\node[itemBox] (item3) at (4.90, -4.60) {};
\node[itemBox] (item4) at (4.90, -6.25) {};

\node[draw=warrantGreen,circle,line width=0.8pt, minimum size=0.45cm,
      text=warrantGreen,font=\bfseries\small,inner sep=0pt] (ico1) at ($(item1.west)+(0.35,0)$) {$\checkmark$};

\node[draw=warnOrange,regular polygon,regular polygon sides=3, line width=0.8pt, 
      minimum size=0.48cm,text=warnOrange,font=\bfseries\scriptsize,inner sep=0pt] (ico2) at ($(item2.west)+(0.35,-0.02)$) {\raisebox{-0.5pt}{!}};

\node[text=warnOrange,font=\large,inner sep=0pt] (ico3) at ($(item3.west)+(0.35,0)$) {$\bowtie$};

\begin{scope}[shift={($(item4.west)+(0.15,-0.24)$)}]
  \draw[fill=attnBlue,draw=none] (0,0) rectangle (0.07,0.18);
  \draw[fill=attnBlue,draw=none] (0.11,0) rectangle (0.18,0.30);
  \draw[fill=attnBlue,draw=none] (0.22,0) rectangle (0.29,0.42);
  \draw[fill=attnBlue,draw=none] (0.33,0) rectangle (0.40,0.55);
\end{scope}
\coordinate (ico4) at ($(item4.west)+(0.60,0)$);

\node[align=left, font=\small, text width=2.9cm, anchor=west] at ($(ico1.east)+(0.15,0)$) {
  \textbf{item 1:}\\[-2pt] {\scriptsize support / true evidence}
};
\node[align=left, font=\small, text width=2.9cm, anchor=west] at ($(ico2.east)+(0.13,0.02)$) {
  \textbf{item 2:}\\[-2pt] {\scriptsize relevant distractor}
};
\node[align=left, font=\small, text width=2.9cm, anchor=west] at ($(ico3.east)+(0.15,0)$) {
  \textbf{item 3:}\\[-2pt] {\scriptsize stale history}
};
\node[align=left, font=\small, text width=2.9cm, anchor=west] at (ico4) {
  \textbf{item 4:}\\[-2pt] {\scriptsize weak but useful signal}
};

\draw[attnArrow] (query.east) to[out=38,in=180]
  node[pos=0.45,above=1pt,font=\scriptsize,text=attnBlue] {$\alpha_{i1}$}
  (item1.west);

\draw[attnArrow] (query.east) to[out=15,in=180]
  node[pos=0.45,above=1pt,font=\scriptsize,text=attnBlue] {$\alpha_{i2}$}
  (item2.west);

\draw[attnArrow] (query.east) to[out=-8,in=180]
  node[pos=0.45,below=1pt,font=\scriptsize,text=attnBlue] {$\alpha_{i3}$}
  (item3.west);

\draw[attnArrow] (query.east) to[out=-28,in=180]
  node[pos=0.45,below=1pt,font=\scriptsize,text=attnBlue] {$\alpha_{i4}$}
  (item4.west);

\node[formulaBoxBlue] at (3.72,-7.65)
  {$\displaystyle h_i=\sum_j \alpha_{ij}v_j$};

\node[align=left,font=\small,text=darkText] at (3.55,-8.75) {
  {\color{attnBlue}$\alpha_{ij}$}: relevance / access\\[-1pt]
  {\color{attnBlue}$c_{ij}=\alpha_{ij}v_j$}: read-derived contribution
};

\node[font=\bfseries\large,text=warnOrange] at (3.72,-9.50)
  {Relevant $\ne$ evidential};

\node[panelTitle, text=warrantGreen] at (7.65, 0) {\textbf{b} \  Warrant Admission};

\node[warrantBox] (cij) at (9.25,-2.60) {
  \textbf{read-derived}\\
  \textbf{contribution}\\[3pt]
  $\displaystyle c_{ij}=\alpha_{ij}v_j$
};

\node[warrantBox] (gij) at (12.35,-2.60) {
  \textbf{query-item}\\
  \textbf{permission}\\[3pt]
  $\displaystyle g_{ij}=g(q_i,k_j)$
};

\node[warrantBox] (cwij) at (15.45,-2.60) {
  \textbf{warranted}\\
  \textbf{evidence}\\
  \textbf{contribution}\\[3pt]
  $\displaystyle c^W_{ij}=\alpha_{ij}g_{ij}v_j$
};

\node[warrantBox,text width=1.9cm] (metric) at (18.55,-2.60) {
  \textbf{metric-facing}\\
  \textbf{prediction}\\
  \textbf{object}\\[3pt]
  {\scriptsize score / logit /}\\[-1pt]
  {\scriptsize state / copy mass}
};

\draw[dashedGrayArrow] (cij.east) --
  node[above=1pt,font=\scriptsize\itshape,text=grayLine,align=center] {path\\[-2pt]exposure}
  (gij.west);

\draw[blackArrow] (gij.east) -- (cwij.west);
\draw[blackArrow] (cwij.east) -- (metric.west);

\node[bigFormulaGreen] at (13.95,-5.10)
  {$\displaystyle h_i^W=\sum_j \alpha_{ij}g_{ij}v_j$};

\node[badge,minimum width=2.55cm] at (10.15,-6.60)
  {\textcolor{warrantGreen}{$\bullet\!\bullet\!\bullet$}\quad item-wise};

\node[badge,minimum width=3.15cm] at (13.42,-6.60)
  {\textcolor{warrantGreen}{$\blacklozenge$}\quad query-conditioned};

\node[badge,minimum width=3.25cm] at (16.85,-6.60)
  {\textcolor{warrantGreen}{$\oslash$}\quad no attention\\[-1pt]renormalization};

\draw[dashedGreen,rounded corners=4pt] (9.60,-8.65) rectangle (18.30,-7.50);
\node[align=left,font=\small] at (13.95,-8.07) {
  \textcolor{grayLine}{\textbf{PathExposureOnly:}}\quad expose path, $g=1$\\[2pt]
  \textcolor{warrantGreen}{\textbf{Full Warrant:}}\quad expose path + learned $g_{ij}$
};

\node[panelTitle, text=grayLine] at (20.45, 0) {\textbf{c} \  Formulation Comparison};

\node[grayCompBox] (glu) at (23.75,-1.80) {
  \textbf{Post-attention GLU}\\[3pt]
  $\displaystyle y_i=F\!\left(\sum_j \alpha_{ij}v_j\right)$
};

\node[align=center,font=\small\itshape,text=grayLine] at (23.75,-3.20) {
  after aggregation\\[-1pt]
  item identity already mixed
};

\node[grayCompBox] (logit) at (23.75,-4.55) {
  \textbf{Attention-logit gate}\\[3pt]
  $\displaystyle \tilde{\alpha}_{ij}
  =\operatorname{softmax}_j(s_{ij}+\log r_{ij})$
};

\node[align=center,font=\small\itshape,text=grayLine] at (23.75,-5.65) {
  changes relevance\\[-1pt]
  redistributes mass
};

\node[grayCompBox] (wg) at (23.75,-7.10) {
  \textbf{Warrant}\\[3pt]
  $\displaystyle y_i=\sum_j\alpha_{ij}g(q_i,k_j)v_j$
};

\node[align=center,font=\small\itshape,text=warrantGreen] at (23.75,-8.40) {
  before aggregation\\[-1pt]
  gates contribution, not relevance
};

\node[align=center,font=\small] at (23.75,-9.37) {
  $\displaystyle
  F\!\left(\sum_j\alpha_{ij}v_j\right)
  \ne
  \sum_j \alpha_{ij}g(q_i,k_j)v_j$
};

\node[panelTitle, text=warrantGreen] at (0, -10.60) {\textbf{d} \  Path Localization Across Domains};

\node[draw=warrantGreen,fill=white,rounded corners=4pt,
      line width=0.8pt,minimum width=7.2cm,minimum height=0.75cm,
      align=center] (op) at (13.52,-12.00)
  {$\displaystyle \alpha_{ij}v_j \;\longrightarrow\;
    \alpha_{ij}g(q_i,k_j)v_j$};

\node[taskBox] (ctdg) at (2.75,-14.20) {
  \hspace{0.5cm}\textbf{CTDG:}\\[-1pt]
  \hspace{0.5cm}history term $\to$\\[-1pt]
  \hspace{0.5cm}edge score
};

\node[taskBox] (mtpp) at (8.11,-14.20) {
  \hspace{0.5cm}\textbf{MTPP:}\\[-1pt]
  \hspace{0.5cm}past event term $\to$\\[-1pt]
  \hspace{0.5cm}mark logit
};

\node[taskBox] (rag) at (13.52,-14.20) {
  \hspace{0.5cm}\textbf{RAG:}\\[-1pt]
  \hspace{0.5cm}passage term $\to$\\[-1pt]
  \hspace{0.5cm}support logit
};

\node[taskBox] (stpp) at (18.93,-14.20) {
  \hspace{0.5cm}\textbf{STPP:}\\[-1pt]
  \hspace{0.5cm}spatial history $\to$\\[-1pt]
  \hspace{0.5cm}location state
};

\node[taskBox] (tkg) at (24.30,-14.20) {
  \hspace{0.5cm}\textbf{TKG:}\\[-1pt]
  \hspace{0.5cm}historical fact $\to$\\[-1pt]
  \hspace{0.5cm}tail score / copy path
};

\coordinate (branch) at (13.52,-12.80);
\draw[dashedGreen] (op.south) -- (branch);
\draw[dashedGreen] (branch) -| (ctdg.north);
\draw[dashedGreen] (branch) -| (mtpp.north);
\draw[dashedGreen] (branch) -- (rag.north);
\draw[dashedGreen] (branch) -| (stpp.north);
\draw[dashedGreen] (branch) -| (tkg.north);

\begin{scope}[shift={($(ctdg.west)+(0.20,0)$)},scale=0.45]
  \draw[attnBlue,line width=1pt] (0,0)--(0.7,0.45)--(1.25,-0.1)--(0.55,-0.55)--cycle;
  \draw[attnBlue,line width=1pt] (0.7,0.45)--(0.55,-0.55);
  \foreach \p in {(0,0),(0.7,0.45),(1.25,-0.1),(0.55,-0.55)}
    \fill[attnBlue] \p circle (0.10);
\end{scope}

\begin{scope}[shift={($(mtpp.west)+(0.20,0)$)},scale=0.42]
  \draw[draw=purple,fill=purple!10,line width=1pt,rounded corners=1pt] (0,-0.6) rectangle (1.2,0.55);
  \draw[purple,line width=1pt] (0,0.25)--(1.2,0.25);
  \draw[purple,line width=1pt] (0.25,0.55)--(0.25,0.78);
  \draw[purple,line width=1pt] (0.95,0.55)--(0.95,0.78);
  \draw[purple,line width=1pt] (1.35,-0.35) circle (0.35);
  \draw[purple,line width=1pt] (1.35,-0.35)--(1.35,-0.12);
  \draw[purple,line width=1pt] (1.35,-0.35)--(1.55,-0.45);
\end{scope}

\begin{scope}[shift={($(rag.west)+(0.25,0)$)},scale=0.45]
  \draw[draw=attnBlue,fill=attnBlue!8,line width=1pt] (0,-0.7) rectangle (0.95,0.7);
  \draw[attnBlue,line width=1pt] (0.18,0.32)--(0.75,0.32);
  \draw[attnBlue,line width=1pt] (0.05,0.02)--(0.75,0.02);
  \draw[attnBlue,line width=1pt] (0.05,-0.28)--(0.60,-0.28);
\end{scope}

\begin{scope}[shift={($(stpp.west)+(0.45,0)$)}]
  \draw[draw=warnOrange, fill=warnOrange!15, line width=1pt] (0,-0.4) ellipse (0.32 and 0.10);
  \draw[draw=warnOrange, fill=warnOrange, line width=0.8pt] (0,-0.4) -- (-0.18, 0.15) arc (210:-30:0.21) -- cycle;
  \fill[white] (0,0.25) circle (0.08);
\end{scope}

\begin{scope}[shift={($(tkg.west)+(0.50,0)$)},scale=0.45]
  \draw[draw=brown,line width=1pt] (0,0.55) ellipse (0.65 and 0.20);
  \draw[draw=brown,line width=1pt] (-0.65,0.55)--(-0.65,-0.65);
  \draw[draw=brown,line width=1pt] (0.65,0.55)--(0.65,-0.65);
  \draw[draw=brown,line width=1pt] (0,-0.65) ellipse (0.65 and 0.20);
  \draw[draw=brown,line width=1pt] (-0.65,0.15) arc (180:360:0.65 and 0.20);
  \draw[draw=brown,line width=1pt] (-0.65,-0.25) arc (180:360:0.65 and 0.20);
\end{scope}

\node[font=\itshape\small,text=darkText] at (13.52,-15.80)
  {Apply Warrant only where the weighted value term reaches the reported metric.};

\end{tikzpicture}

}
\captionof{figure}{Implementation overview. Attention-derived item contributions are connected to the prediction interface and controlled by permission. Path construction is the implementation procedure for applying and evaluating permission.}
\label{fig:main_message}
\end{center}

\subsection{Placement and Adapter Controls}
\label{subsec:path_localization_adaptation}

To check the implementation of the permission interface, we compare prediction-facing placement with generic q--k placement and shuffled pairing. Figure~\ref{fig:path_localization} reports limited-budget diagnostics, while Table~\ref{tab:path_placement_controls} reports the main-budget comparison on LastFM/DyGFormer. Comparisons across different paths measure changes in the entire connection configuration, rather than permission alone.

\begin{center}
\includegraphics[width=0.82\textwidth]{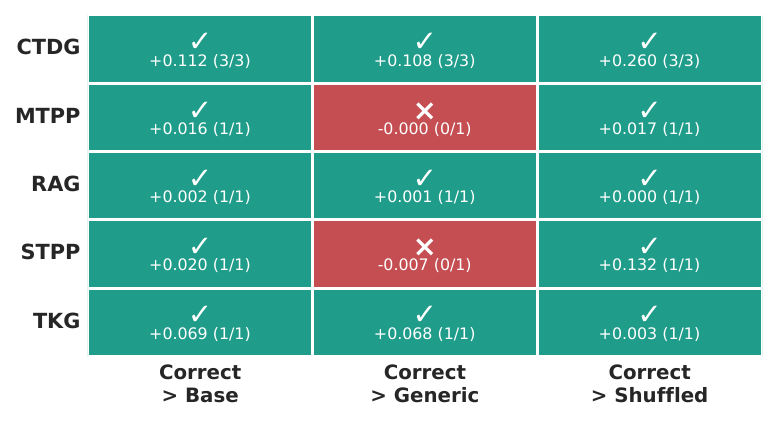}
\captionof{figure}{Path localization check. Each domain compares Base, Generic q-k \method{}, Correct-path \method{}, and Shuffled pairing.}
\label{fig:path_localization}
\end{center}

Correct-path \method{} outperforms Base in the improvement direction of each domain's metric. On CTDG and TKG, Generic q-k \method{} is nearly flat, whereas Correct-path \method{} yields large gains. Generic obtains +0.0043 AUC on CTDG versus +0.1119 for Correct-path, and +0.0004 MRR on TKG versus +0.0687 for Correct-path. These values include both path construction and permission, rather than measuring permission's additional effect.

Shuffled pairing shows that query--item correspondence matters for prediction, but does not attribute that effect to the gate alone. In CTDG, the edge-query adapter and opened path are also affected. Figure~\ref{fig:path_localization} therefore provides supporting evidence for an interface that preserves query--item correspondence; permission's additional effect is evaluated in Section~\ref{subsec:relevance_permission_separation} and Table~\ref{tab:cross_domain_path_permission}.

\begin{center}
\begin{minipage}{\textwidth}
\centering
\small
\captionof{table}{Placement controls on LastFM/DyGFormer. Generic q--k placement and the prediction-facing path are compared under the same main-budget, three-seed conditions. Differences between placements include connection configuration, not permission alone.}
\label{tab:path_placement_controls}
\vspace{2mm}
\begin{tabular}{lc}
\toprule
CTDG LastFM / DyGFormer variant & AUC \\
\midrule
Base & .8482 \\
Generic q--k permission only & .8484$\pm$.0074 \\
Localized OpenPath & .9010 \\
Generic permission + Localized OpenPath & .9022$\pm$.0012 \\
Localized Full & .9023 \\
Generic permission + Localized Full & \textbf{.9040$\pm$.0023} \\
\bottomrule
\end{tabular}
\end{minipage}
\end{center}

Table~\ref{tab:path_placement_controls} directly separates placement and path exposure under the same LastFM/DyGFormer main-budget setting. Generic q--k permission obtains .8484, close to Base at .8482, whereas localized OpenPath reaches .9010. This difference arises from localization and exposure of the metric-facing contribution path. Full Warrant applies permission at this interface, with its conditional marginal effect measured in Section~\ref{subsec:relevance_permission_separation} and Table~\ref{tab:cross_domain_path_permission}.

\subsubsection{CTDG Edge-Query Adapter Ablation}
\label{subsec:edge_query_ablation}

In CTDG, the prediction target is the current source--destination edge score rather than a node representation. The \method{} query must therefore represent an edge-level prediction request. The edge-query adapter ablation tests whether this interface is explained by additional parameters or handcrafted temporal features. Figure~\ref{fig:edge_query_ablation} summarizes the results.

\begin{center}
\includegraphics[width=0.92\textwidth]{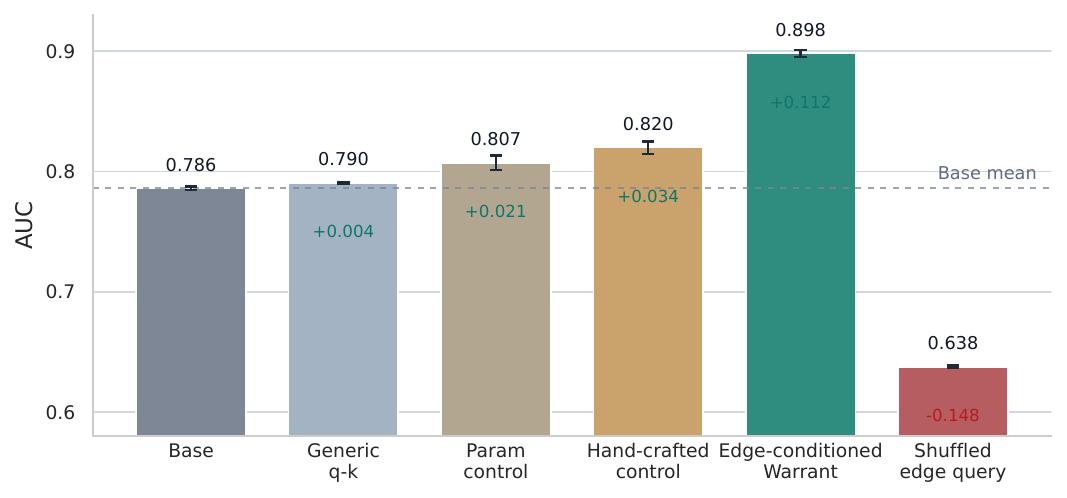}
\captionof{figure}{CTDG edge-query adapter ablation. Edge-conditioned \method{} outperforms generic \method{}, parameter controls, and handcrafted controls, while shuffled edge queries cause a large performance drop.}
\label{fig:edge_query_ablation}
\end{center}

Edge-conditioned \method{} obtains AUC 0.8980, exceeding Base by +0.1119. Parameter and handcrafted-only controls also improve performance, but less than Edge-conditioned \method{}. Breaking the pairing between edge queries and history items at evaluation reduces AUC to 0.6376. Together with the main-budget OpenPath results, these observations locate the main CTDG changes in metric-facing path construction and query alignment for the current edge prediction.

\end{document}